\documentclass[letterpaper]{article} %
\usepackage{aaai2026}  %
\usepackage{times}  %
\usepackage{helvet}  %
\usepackage{courier}  %
\usepackage[hyphens]{url}  %
\usepackage{graphicx} %
\usepackage{natbib}  %
\usepackage{caption} %
\usepackage{algorithm}
\usepackage{algorithmic}
\usepackage{enumitem}
\usepackage{tabularx}
\usepackage{booktabs}
\usepackage{multirow}
\usepackage{colortbl}
\usepackage{amsmath}   
\usepackage{amssymb}   
\usepackage{rotating}
\usepackage{xcolor}
\usepackage{adjustbox}
\usepackage{placeins}

\usepackage{amsthm}
\theoremstyle{plain}
\newtheorem{theorem}{Theorem}[section]

\theoremstyle{definition}
\newtheorem{definition}[theorem]{Definition}

\theoremstyle{remark}

\usepackage{newfloat}
\usepackage{listings}
\DeclareCaptionStyle{ruled}{labelfont=normalfont,labelsep=colon,strut=off} %
\floatstyle{ruled}
\newfloat{listing}{tb}{lst}{}
\floatname{listing}{Listing}
\title{A Novel Approach to Evaluating Evaluation Metrics for Multi-Output Structured Prediction}
\author{
    Akshay Vyas,
    Angelo Pimienta,
    Nicholas Ruozzi
}
\affiliations{
    The University of Texas at Dallas\\    
    akshay.vyas, angelomatthew.pimienta, nicholas.ruozzi@utdallas.edu
}

\usepackage{bibentry}

\begin{document}

\maketitle
\begin{abstract}
In multi-output structured prediction tasks, while only one ground truth label may be provided in the training data, multiple equally valid outputs may be possible, making reliable evaluation a persistent challenge. We postulate that human evaluators implicitly use task-specific invariants, e.g., object boundaries in colorized images or named entities in translations, to judge if an output is acceptable. Under this assumption, we introduce a notion of approximate task-specific invariants and use them as diagnostic tools to evaluate a variety of existing metrics for vision and language tasks. We use these task invariants as part of a framework to systematically test metric reliability by encouraging domain-relevant invariants in model outputs via an augmented loss function. In our experiments, we observe that enforcing invariants with an augmented loss yields substantial improvements in popular distributional metrics while more traditional metrics change only marginally. Through this invariants-driven evaluation, we expose where standard metrics fail to detect meaningful differences, and we highlight the conditions under which distributional metrics succeed or still fall short.    
\end{abstract}
\begin{links}
    \link{Code}{https://github.com/akshay25vyas/invariants-metric-eval}
    \link{Proceedings version}{https://doi.org/10.1609/aaai.v40i31.39808}
\end{links}

\section{Introduction}
Machine learning models trained for multi-output vision and language tasks often face an evaluation paradox: there may exist many plausible outputs for a given input, e.g., different colorizations of a photo or multiple valid translations of a sentence, yet traditional evaluation metrics assume a single ground-truth output. In vision, pixel-level metrics such as mean squared error (MSE) or peak signal-to-noise ratio (PSNR) treat any deviation from the single ground-truth image as an error, even if the deviation is a valid alternative, e.g., a shirt colored blue instead of green. These metrics can also be poor proxies for perceptual quality, e.g., an image can have low MSE yet look unnatural to humans. In language, n-gram overlap metrics such as Bilingual Evaluation Understudy Score (BLEU), while standard, can be brittle: a translation that uses synonyms or different phrasing will get a low BLEU score despite conveying the same meaning.
 
Overall, when trained models output one of many acceptable answers, the above metrics can often over penalize legitimate variation while under penalizing more human perceptible errors, which can lead trained models to produce labels that may have low error with respect to the chosen metric but be easily identified as poor by human evaluators.

Recently, distributional/learned metrics have emerged as attempts to address the above issues. Distributional metrics compare the distribution of labels produced by a given model to the distribution of the ground truth labels. For images, the Fréchet Inception Distance (FID) \citep{heusel2017gans} computes the Wasserstein-2 distance between multivariate normal distributions fit to model generated labels and ground-truth labels in a feature space extracted from the Inception network.  As only distributions are considered, these metrics allow for diversity beyond pixel agreement. For text, BERTScore leverages pre-trained language model embeddings to measure semantic similarity between the model-generated text and ground-truth reference. Distributional metrics can better handle paraphrasing or varying image styles, allowing them to be less influenced by superficial differences between the predicted and the ground-truth labels. However, an open question remains: Do these distributional metrics capture the variations that human evaluators expect, or do they have blind spots of their own? For example, if a translation preserves meaning but shifts tone, will Fréchet BERT Distance (FBD) register the change? Does FID drop when an image's colors violate object boundaries, or could it be fooled by overall color statistics? These questions highlight the need for a systematic study of metric sensitivity to meaningful variations in output.

The main postulate in this work is that human evaluators implicitly or explicitly apply domain or task invariants when judging neural network outputs. Invariants are properties that should remain (approximately) unchanged under valid transformations of the output. For example, a human evaluator for the image colorization task might expect that the addition of color does not alter the image's structural details, e.g., the model colorized image should respect the same object boundaries and depth cues as the original ground-truth color image. If neural network colorization causes ``color bleeding'' across object edges, humans might immediately flag the colorization as incorrect, even if the colors appear plausible overall. Similarly, when assessing machine translation, bilingual speakers might expect the translation to preserve named entities and sentiment, e.g., a translated sentence that flips the sentiment or mistranslates a proper name violates one of these invariants. In image captioning, an accurate caption should reflect the image's content and relationships, e.g., who is doing what to whom. So, properties like semantic roles and scene relationships act as invariants, e.g., if the image shows a cat on a couch, predicted captions should maintain that subject-object relation. 

While the above invariants are natural for their respective domains, many existing metrics do not explicitly account for these kinds of invariants. Further, existing work has observed that neural networks trained for structured prediction tasks do not necessarily produce predictions that enforce the task constraints, e.g., neural networks trained to predict shortest paths need not produce a connected path, a natural invariant for path problems \citep{ahmed2022}.  We observe the same difficulties for the multi-output structured prediction tasks described above: task invariants are not automatically preserved through neural network training (even when training with large data sets).  However, as violation of these invariants can result in poor-quality labels that are readily identified as such by human evaluators, our aim is to understand to what extent natural task invariants are captured by existing metrics.  
We propose a framework to evaluate the evaluation metrics themselves using approximate invariants as a probe. Rather than artificially introducing invariant violations, we systematically compare outputs from standard baseline models (which can significantly violate the task-specific invariants) with invariant-regularized models explicitly trained to respect approximate invariants, e.g., segmentation consistency in image colorization, named entities in translation, etc. 
We consider three modeling tasks: image colorization, machine translation, and image captioning.  For each task, we first define one or more approximate task invariants (see Table 1 for a list of possible invariants). We then train models that are both invariant aware, where we enforce the invariants with an auxiliary loss function, and invariant agnostic. We evaluate the different models via a battery of metrics including standard metrics and distributional metrics and observe how the performance changes when the invariant is or is not enforced.  
 Our contributions are as follows.

\begin{itemize}[leftmargin = .2in]
\item We introduce a novel evaluation framework that uses approximate task invariants as a diagnostic lens for evaluating metrics that provides an alternative to human evaluators. 
\item We apply our framework across vision and language domains (image colorization, image captioning, and machine translation) with distinct invariants and evaluation challenges. This comprehensive study demonstrates the versatility of our approach and provides insights into metrics ranging from pixel-level (MSE) and token-level (BLEU) to distributional (FID) and embedding-based (BERTScore).
\item Our experiments show that distributional metrics, e.g., FID, FBD, etc., are generally more sensitive to invariant-related differences, and enforcing invariants in outputs can lead to significant performance gains ($15\%-30\%$) in sensitive metrics. In constrast, we find that traditional metrics, e.g., BLEU, PSNR, etc., are largely unchanged when training with the selected invariant losses. 
\end{itemize}

\begin{table}[!t]
\scriptsize
\centering
\begin{tabularx}{\linewidth}{@{}lX@{}} 
\toprule
\textbf{Main Task} & \textbf{Invariant Task(s)} \\ \midrule
\textbf{Image Colorization} &  Image Segmentation, Depth Estimation, Edge Detection \\ 
\hline
\textbf{Machine Translation (NMT)} &  Named Entity Recognition (NER), Sentiment Analysis, Paraphrase Detection \\ 
\hline
\textbf{Image Captioning} &   Semantic Role Labeling (SRL), Scene Graph Consistency, Named Entity Recognition (NER) \\ 
\hline
Text Summarization &  Named Entity Recognition (NER), Sentiment Analysis, Coreference Resolution \\ 
\hline
Video Prediction &  Optical Flow Estimation, Object Tracking, Action Recognition \\ 
\hline
Natural Language Inference &  
Paraphrase detection, Sentiment Analysis \\

 \bottomrule
\end{tabularx}
\caption{Tasks and corresponding invariants. {BOLD} tasks are considered in this work.}

\label{tab:diverse_valid_outputs_invariant}

\end{table}

\section{Related Work}

While the aim in this work is to provide alternatives to human evaluators, one branch of related work focuses on metrics that better align with human judgment by incorporating semantic information or learning directly from human evaluations. METEOR \citep{banerjee2005meteor} extended BLEU by accounting for synonyms and stemming, CIDEr \citep{vedantam2015cider} reweights n-grams by TF-IDF to reflect their salience, and SPICE \citep{anderson2016spice} incorporates semantic propositional knowledge. %
SPICE was shown to achieve a higher correlation with human caption preferences than BLEU or CIDEr \citep{anderson2016spice}. \citet{hasler2003measuring} introduced Colorfulness to measure vibrancy in images. 

Learned metrics based on neural network embeddings have also gained traction. BERTScore \citep{zhang2020bertscore} leverages pretrained language model embeddings to judge text similarity, yielding significantly improved correlation with human assessments of translation quality. In vision, the Learned Perceptual Image Patch Similarity (LPIPS) \cite{zhang2018unreasonable} metric measures image differences in a deep feature space.
These learned or ``feature-based'' metrics effectively compute distances in a feature space extracted from trained models on large datasets, making them sensitive to high-level features. 

Distributional metrics evaluate the distribution of outputs as a whole, rather than scoring individual output-reference pairs. This approach arose prominently in image generation research. For example, FID \citep{heusel2017gans} has become a popular approach for assessing the realism of images produced by GANs. For language tasks, \citet{xiang2021assessing} introduce Fréchet Bert Distance (FBD) 
  
and argue that it better aligns high human judgment than BLEU, METEOR, and ROUGE-L. 
 
Distributional metrics also have important limitations. \citet{chong2020} showed that FID is a biased estimator and proposed a corrective fix. \citet{parmar2021}
showed that compression and resizing have a significant effect on FID. \citet{jayasumana2024} show that FID is sensitive to complex image distortions and does not behave monotonically as the distortions increase.  They propose using maximum mean discrepancy (MMD) calculated over CLIP embeddings instead as MMD is unbiased, CLIP embeddings are richer than Inception features, and MMD better aligns with human evaluators in their experiments.

Problem or domain invariants have historically been used to improve generalization, e.g., data augmentation for rotation equivariance in computer vision tasks. Some neural architectures are built on invariants, e.g., convolutional layers enforce translation equivariance symmetry, and further research has aimed at hard-coding other symmetries, such as rotation equivariance, into convolutional neural networks \cite{cohen2016group,worrall2017harmonic}.  However, to the best of our knowledge, the more complex, approximate invariants, including the types we formally define herein, have not been treated in any systematic way. 

Another related line of work has explored enforcing certain desired properties through multi-task learning or auxiliary losses.  In image colorization, researchers have added auxiliary segmentation or depth prediction tasks to the model so that the colorization network jointly learns to respect object boundaries or 3D structure. Similarly, in machine translation, one might regularize a model to preserve named entities or sentiment by jointly training on a related classification task. These auxiliary loss approaches, e.g., \citep{misra2016cross, ruder2017overview}, can act as a form of regularization. However, the approaches in prior work require extra supervision, e.g., segmentation labels, or complex multi-task training setups.  Contrast this with the approach taken here in which no additional labels are needed -- only access to a pretrained DNN for a desired invariant is required. There has also been progress in explicitly modeling the multi-modal nature of outputs, e.g., using probabilistic frameworks or adversarial generative models that can produce multiple different valid outputs \citep{zhu2017toward, zhao2018bias}. These methods address the inherent ambiguity by aiming to capture the full distribution of outputs, but they often rely on adversarial training and do not guarantee that specific semantic properties are preserved in each output.

\section{Multi-Output Tasks and Invariants}
\label{sec:prelim}
We consider a variant of the standard supervised structured learning task: for data observations $x^{(1)}, \ldots, x^{(M)} \in\mathcal{X}$ our aim is to produce corresponding labels $y^{(1)}, \ldots, y^{(M)}$ that are assumed to be taken from a (structured) set $\mathcal{Y}$.  However, we do not assume that there is a one-to-one mapping between the data observations and the labels.  Specifically, we assume that there exists a function $f:\mathcal{X} \rightarrow 2^\mathcal{Y}$ such that for all $m\in\{1,\ldots, M\}$, $y^{(m)} \in f(x^{(m)})$.  Note that if $f$ maps each element of $\mathcal{X}$ to a unique label, i.e., $|f(x)| = 1$, then this is equivalent to a standard supervised learning task whose aim is to find $f$, but more generally, there could exist $x\in\mathcal{X}$ with $|f(x)| > 1$.  In the general case, our aim will be to produce \textbf{one} valid labeling for each given data observation (note that one could also consider fitting distributions over possible labels, but we do not consider that task here).

As an example, consider the image colorization task which seeks to transform grayscale images into colored variants. Inherently, the task is ill-posed: a single grayscale image can be mapped to multiple, plausible colorized images.  In this task, images are often represented in CIELAB color space as a triple $(L,a,b)$. To map this problem into the above framework, we can choose $\mathcal{X}$ to be a set of grayscale images in Lab color space, $\mathcal{Y}$ to be a set of color images in Lab color space, and $f$ can be defined as the function that maps grayscale images to a set of valid/realistic colorizations. Similarly, consider the task of captioning images:  We can define $\mathcal{X}$ as the set of all uncaptioned real-world images, $\mathcal{Y}$ to be a set of captions, and $f$ can be defined as the function that takes an uncaptioned image to a set of valid captions. %

In both of the above examples, there is not necessarily a unique label $y\in\mathcal{Y}$ for every $x\in\mathcal{X}$, nor is it clear that there is even a notion of a ``best'' label $y\in\mathcal{Y}$ as ``best'' in both of these instances could be subjective.  This creates difficulty when using standard approaches to learning/evaluation in these settings:  for a given $x\in\mathcal{X}$, there could be $y,y'\in\mathcal{Y}$ with $y\neq y'$ that are both ``good'' label choices, but they could be far apart from each other under typical notions of distance, e.g., mean squared error, on $\mathcal{Y}$.

\subsection{Approximate Task Invariants}
\label{sec:inv}

Consider the image colorization task for which a DNN is trained to produce a colored image in Lab space given a grayscale image.  This DNN, when applied to an unseen test image, say of a pedestrian crossing the street, may actually produce a ``good'' colorization, but that colorization may be evaluated poorly under a squared error metric if the colors deviate significantly from the ground truth, e.g., the pedestrian's shirt is colored red by the DNN in contrast to blue in the ground truth.
Different colorizers may produce different colors for the pedestrian's shirt, and unless the pedestrian's shirt is somehow unique, even a human evaluator might not know which color is correct.  However, if a particular colorizer assigns a color to the pedestrian's shirt that then bleeds into the neighboring pixels, then a human evaluator may observe that this colorizer has made a mistake.   

In the above example, the human evaluator only noticed the error because an unexpected artifact was observed.  We can view this as a segmentation error - colors should not spill across object boundaries, or put another way, we expect the segmentations of the ground truth and the colorizer-produced image to (approximately) match.  We call this observation that the outputs of ``good'' colorizers should have the same segmentations as the ground-truth images an \textit{approximate task invariant} for the colorization problem.  
Formally, we will define invariants as follows.

\begin{definition}
Let $g:\mathcal{Y} \rightarrow \mathcal{Z}$. A labeling problem, as defined above, is \textbf{$g$-invariant} if for all $x\in\mathcal{X}$ and all $y,y'\in f(x)$, $g(y) = g(y')$, i.e., for all data observations $x$, $g$ applied to any valid labeling of the data observation $x$ yields the same result.    
\end{definition}

We will typically consider $\mathcal{Z}$ to be a subset of $\mathbb{R}^n$ for some $n$.  Ideally, $\mathcal{Z}$ should be a space on which similarities/differences are easier to detect than on $\mathcal{Y}$.  From the definition of $g$-invariant, we have that if $||g(y) - g(y')|| > 0$, then there does not exist an $x\in\mathcal{X}$ such that $y,y'\in f(x)$, i.e., we can use $||g(y) - g(y')||$ as a surrogate measure for how close $y'$ is to being in the set $f(x)$. Not all invariants, as defined above, are useful for classification: constant functions, e.g., $g(y) = 0$, trivially satisfy this definition. Note also that the definition makes no restrictions on $y\notin f(x)$.  That is, the invariant is a property of all ``good'' labels, but some labels that are not ``good'' can map to the same element of $\mathcal{Z}$.  As a result, these invariants should be thought of as desirable properties for ``good'' labelings, but they place no restrictions on invalid labels.  However, the invariants that will be most useful in practice should distinguish ``good'' labels from ``poor'' labels as much as possible.

The intuition behind considering invariants of this form is that they can act like regularizers.  That is, suppose we trained a DNN model for a labeling task, and on input $x$ with ground truth label $y\in f(x)$ it produces the label $y'$. If $y\neq y'$, the output DNN should still be considered good if $y'\in f(x)$ or if $y'$ is at least close to some member of $f(x)$.   
If the labeling task is $g$-invariant, then we can add an auxiliary loss to the objective to penalize model weights for which $||g(y) - g(y')||$ is large.  This has two benefits: (1) it allows some deviation from $||y - y'||$ in order to better match the specified invariant, and (2) even if $||y - y'||$ is small, the penalty encourages the model to produce labels that approximately preserve the invariant, which hopefully increases the realism of the resulting predictions.

While true invariants might be difficult to obtain, we can still apply this approach with approximate invariants:
\begin{definition}
Let $g:\mathcal{Y} \rightarrow \mathcal{Z}$.  A labeling problem, as defined above, is an \textbf{$\epsilon$-approximate $g$-invariant} for some $\epsilon >0$ if for all $x\in\mathcal{X}$ and all $y,y'\in f(x)$, $||g(y) - g(y')||\leq \epsilon$ for an appropriately chosen norm on $\mathcal{Z}$, i.e., for all data observations $x$, $g$ applied to any valid labeling of the data observation $x$ yields \textit{almost} the same result. 
\end{definition}

Consider again the problem of image colorization.  A reasonable (approximate) invariant for this task is that ``good'' colorizations should be almost indistinguishable to a well-trained image segmentor.  Similarly, good colorizations of the same gray scale images should have similar depth maps.  In the invariant framework above, a DNN trained for image segmentation or depth estimation, while not a perfect invariant for the colorization task as neither is likely to be error free, can be used to define a function $g$ such that the colorization task is approximately $g$-invariant. This is the approach that we take in this work.  

\subsection{Invariants as Regularization}

Existing work has observed that enforcing task constraints can yield noticeable performance improvements in practice \citep{ahmed2022}.  Motivated by this observation, our aim, then, is (1) to understand to what extent natural task invariants are preserved by/reflected in existing metrics and (2) to determine whether or not enforcing task invariants has any performance benefits for multi-output structured prediction tasks. To encourage models to better align with invariants, we take a common approach:  we add an additional penalty to the loss function.

Assume that we are fitting a DNN, represented by $f'_\theta$, which takes as input a data observation $x$ and attempts to produce the ground truth label $y$. We also assume that we are given an approximate invariant $g$ for this labeling task. With this notation, our approach implements an augmented loss function, $L$, given by
\begin{align*}
L(\theta|x,y) = L_T(f'_{\theta}(x),y) + \alpha L_I(g(f'_{\theta}(x)), g(y)),
\end{align*}
where $L_T$ is a task-specific loss measuring the error between the DNN's prediction and the ground truth label $y$, $\theta$ is the vector of parameters for the DNN, $L_I$ is the invariant loss for invariant $g$, measuring the error between the invariant map applied to the ground truth label $y$ and the invariant map applied to the output of the DNN, and $\alpha$ is a hyperparameter controlling the trade-off between these two losses. Note that the invariant $g$ is treated as \textbf{fixed}, and although it will be chosen to be a DNN in our formulation, its parameters are frozen, i.e., they are not learned as part of the framework. As an example, for image colorization with a segmentation invariant, we select $g$ to be a frozen semantic segmentation network and $L_I$ to be the cross-entropy between the segmentation masks of the generated and ground-truth color images.

\section{Experimental Evaluation}

\begin{figure*}[t!]
  \centering
  \setlength{\tabcolsep}{2pt}
  \renewcommand{\arraystretch}{0}
  \resizebox{\textwidth}{!}{%
    \begin{tabular}{c ccc@{\hspace{10pt}}ccc}
      & & \Large Segmentation Regularized & \Large Baseline 
        & & \Large Depth Regularized    & \Large Baseline      \\
      & \Large Ground Truth           & \Large ChromaGAN + Seg 
        & \Large ChromaGAN 
        & \Large Ground Truth           & \Large ColorFormer + Depth 
        & \Large ColorFormer   \\
      \raisebox{1cm}{\rotatebox{90}{\Large Colorized outputs}}\hspace{2mm} &
      \includegraphics[height=5cm,keepaspectratio]{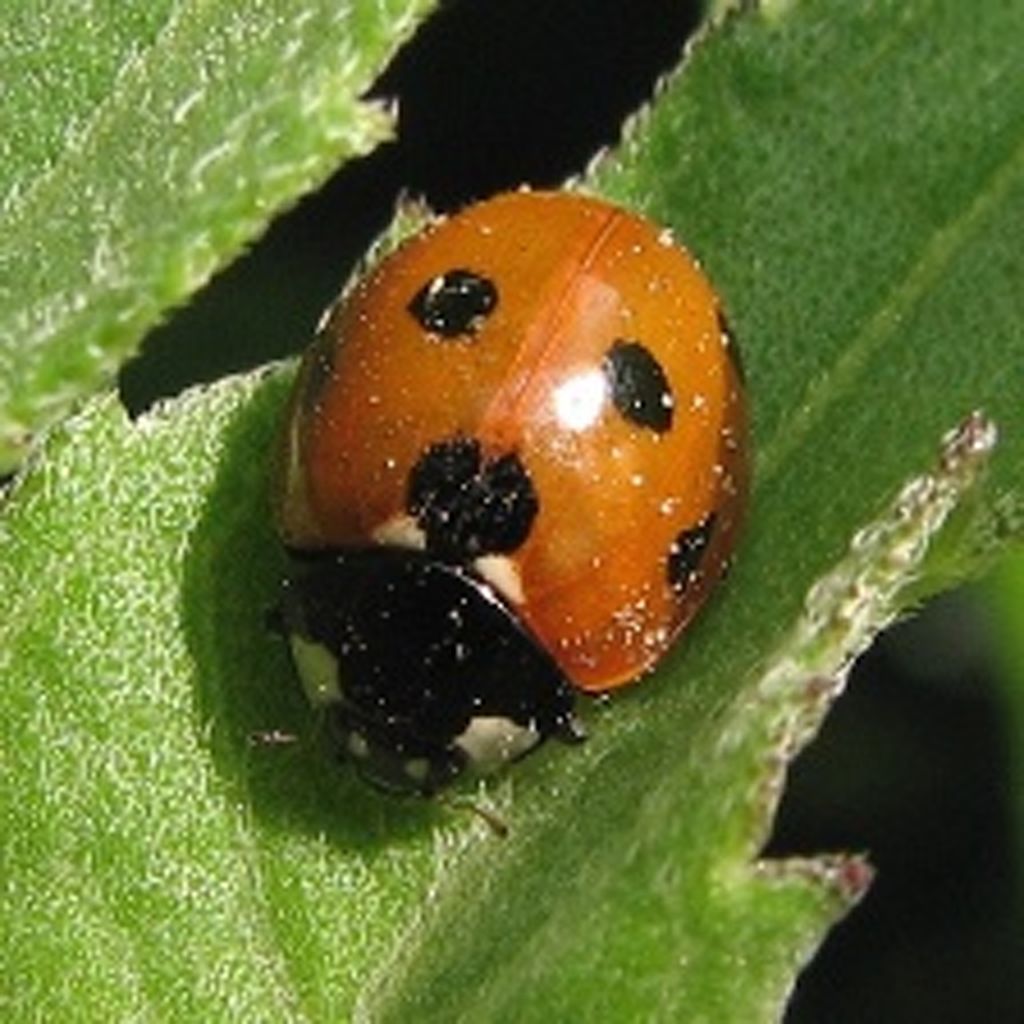}  & 
      \includegraphics[height=5cm,keepaspectratio]{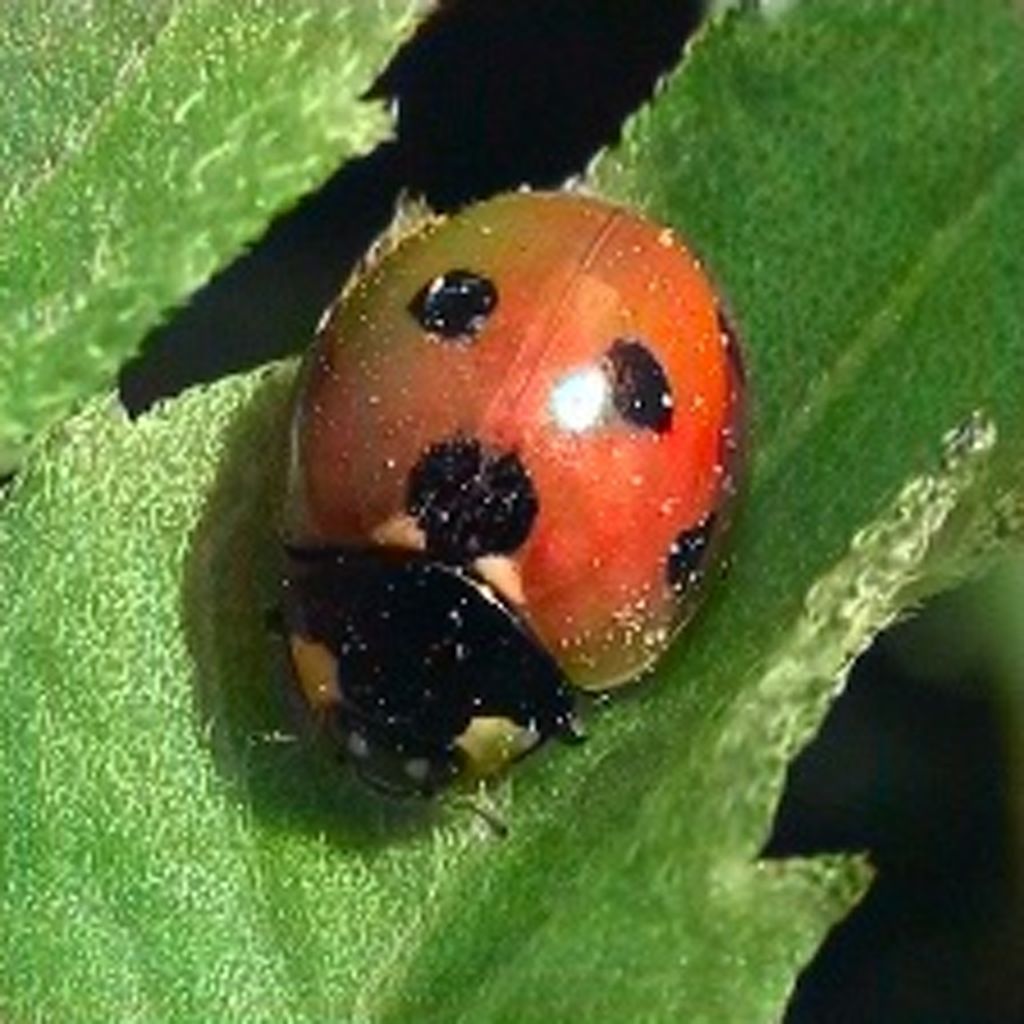}  & 
      \includegraphics[height=5cm,keepaspectratio]{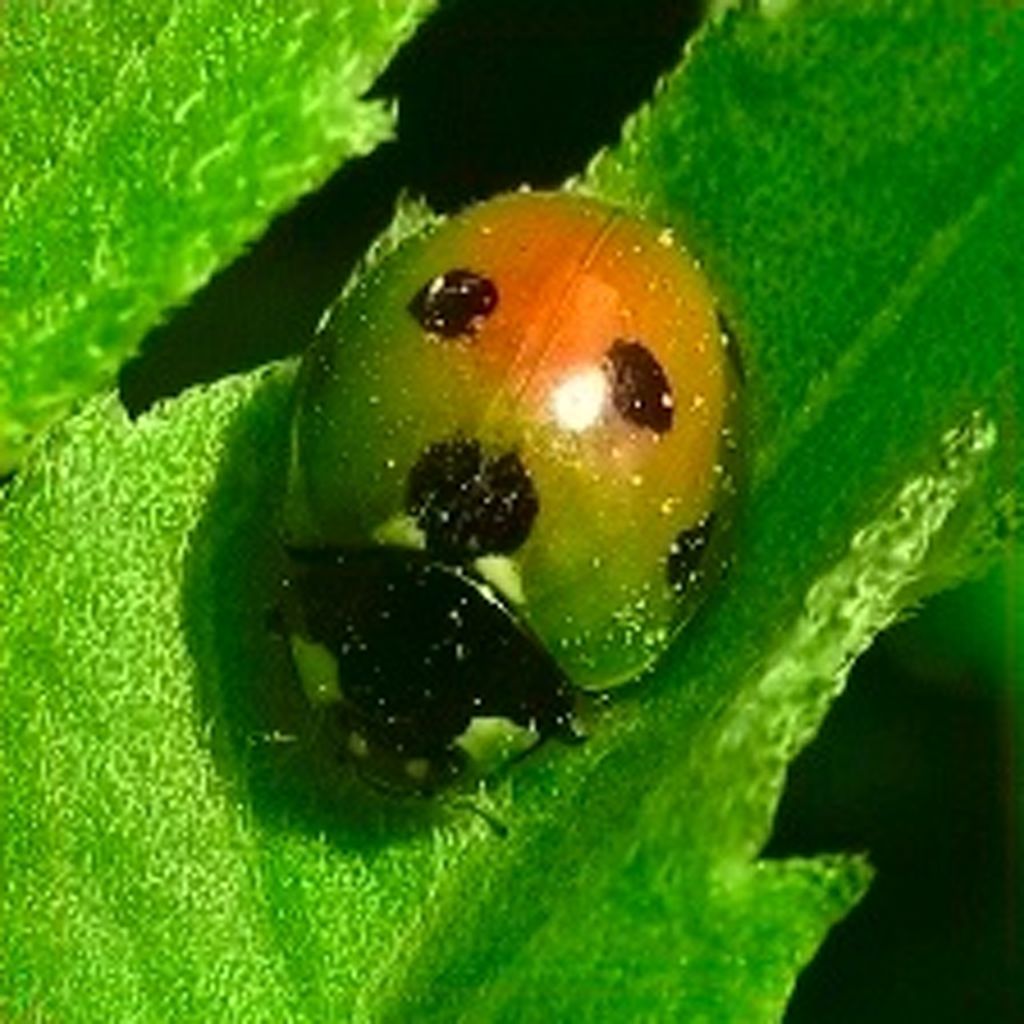}  &
      \includegraphics[height=5cm,keepaspectratio]{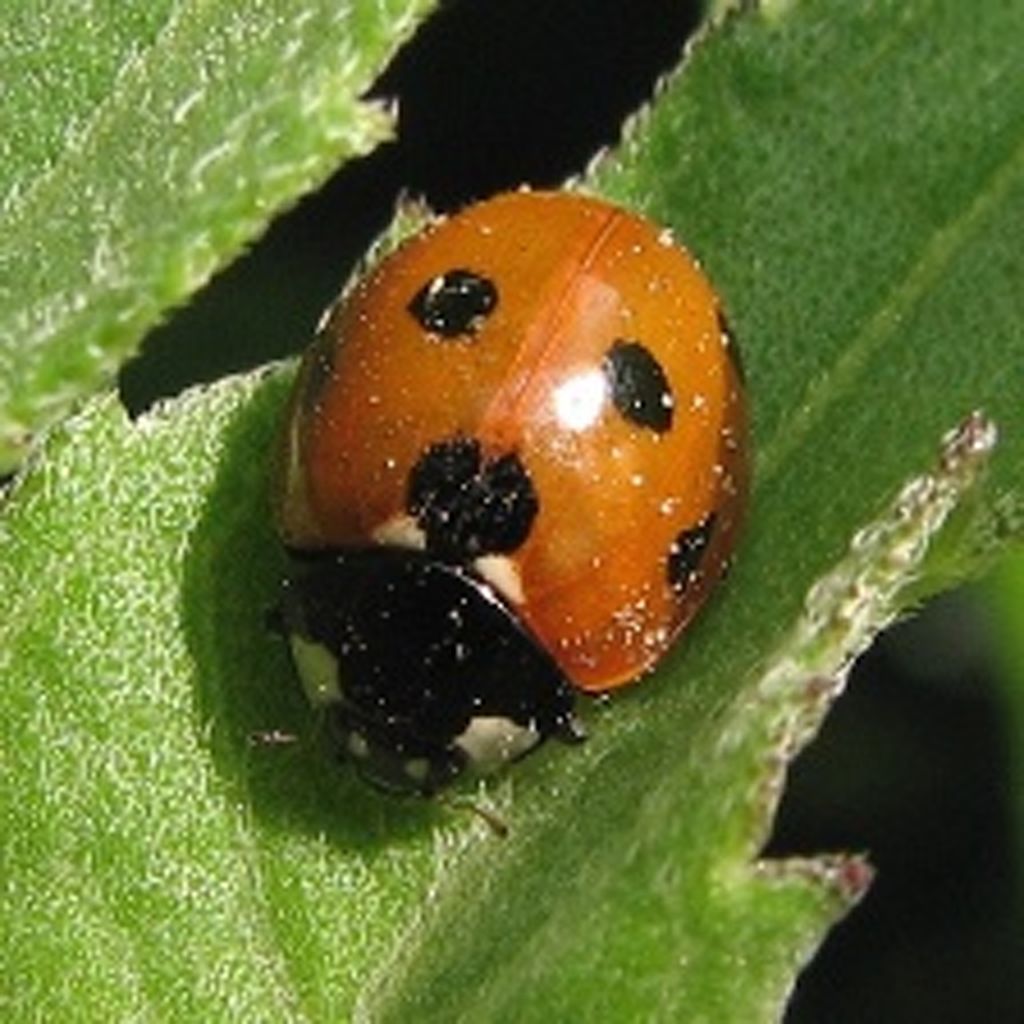}  & 
      \includegraphics[height=5cm,keepaspectratio]{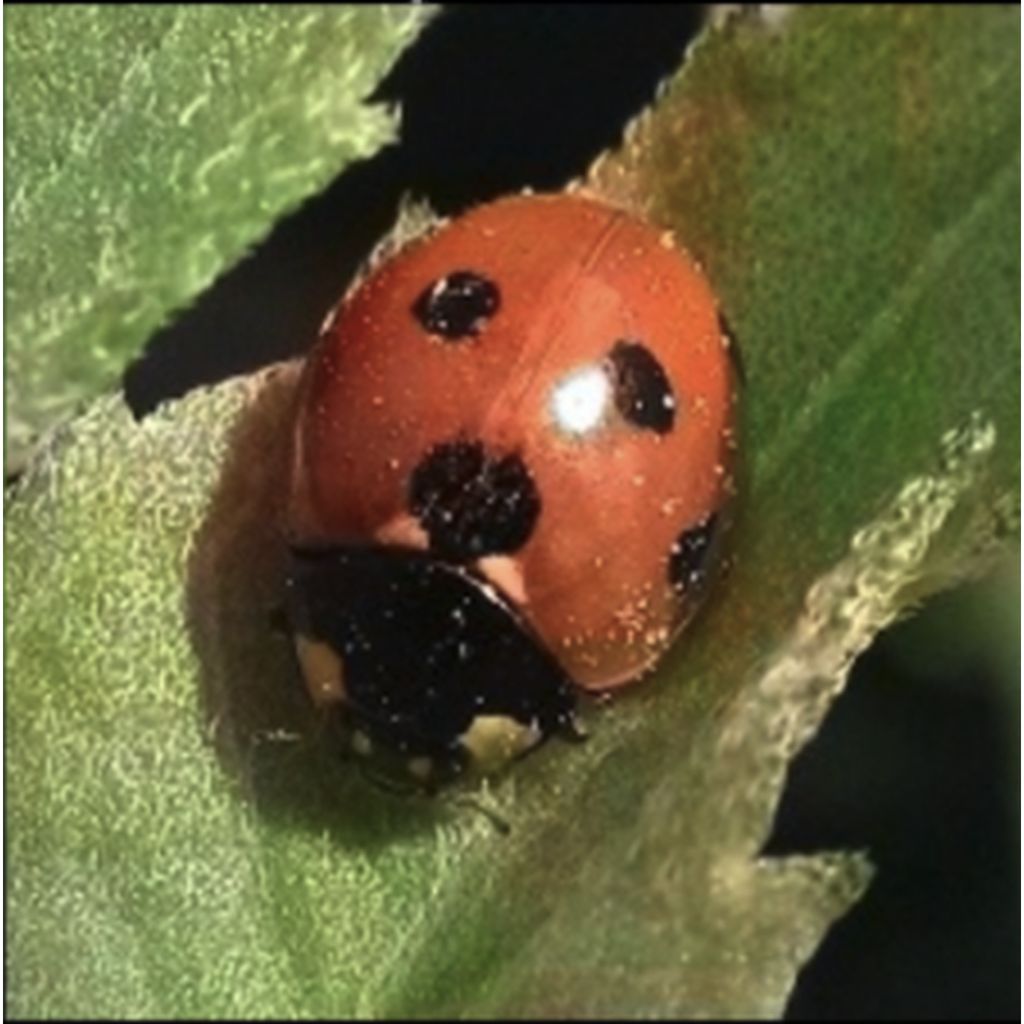}  & 
      \includegraphics[height=5cm,keepaspectratio]{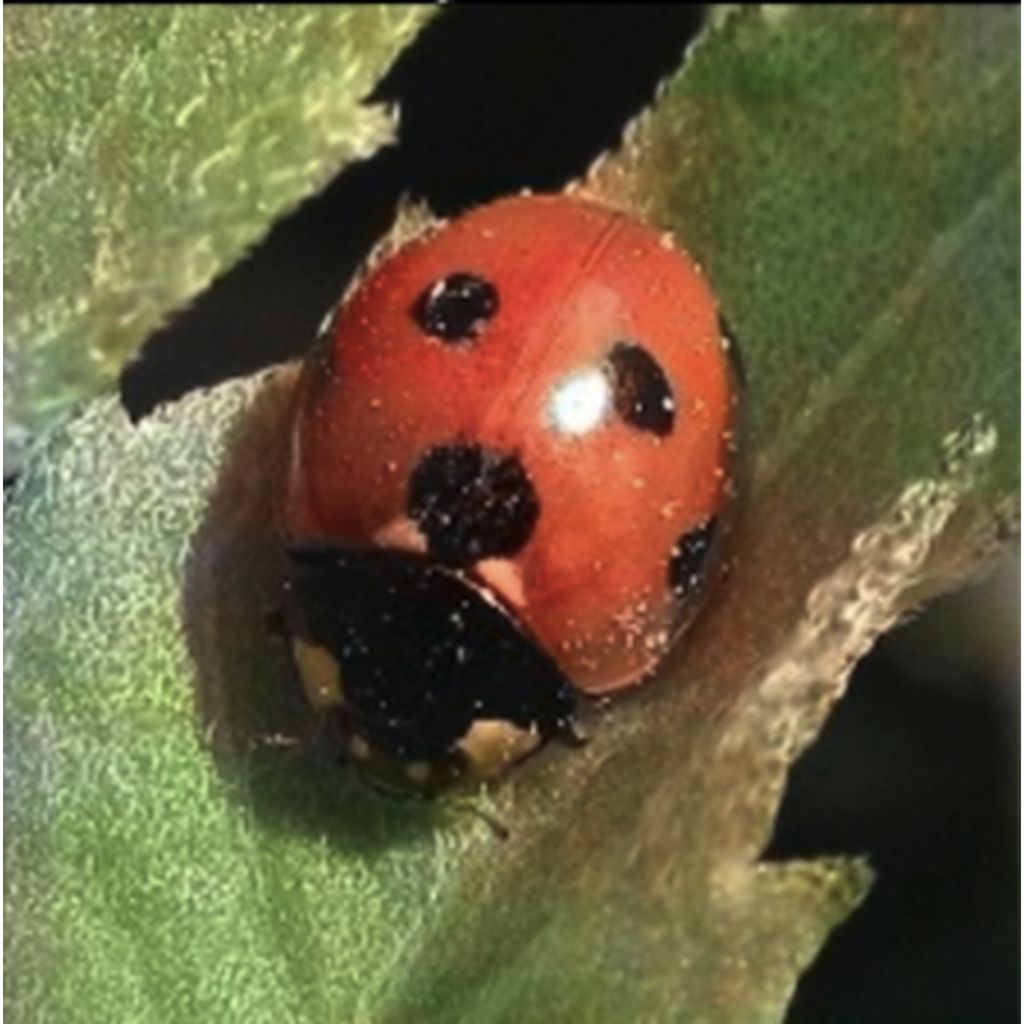}  \\
      [4pt]
      \raisebox{1cm}{\rotatebox{90}{\Large Invariant masks}}\hspace{2mm} &
      \includegraphics[height=5cm,keepaspectratio]{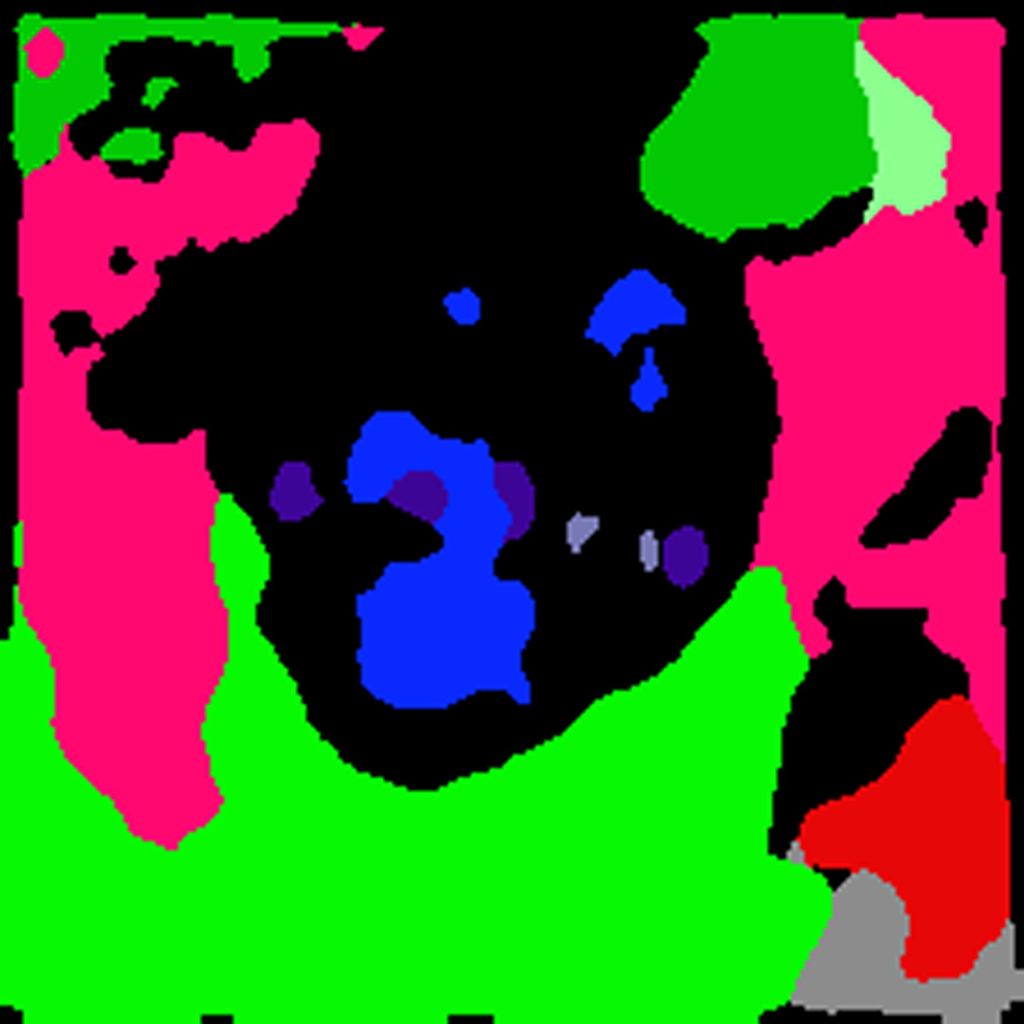}  & 
      \includegraphics[height=5cm,keepaspectratio]{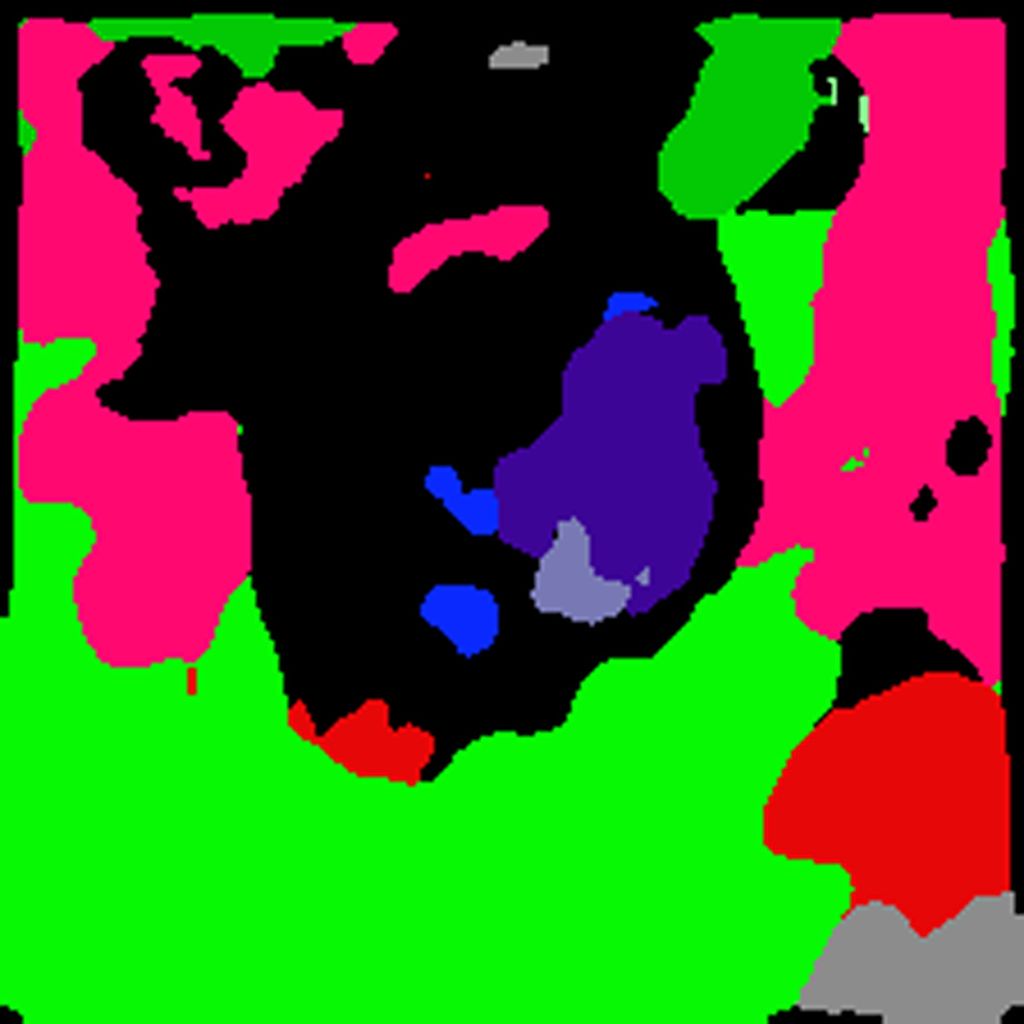}  & 
      \includegraphics[height=5cm,keepaspectratio]{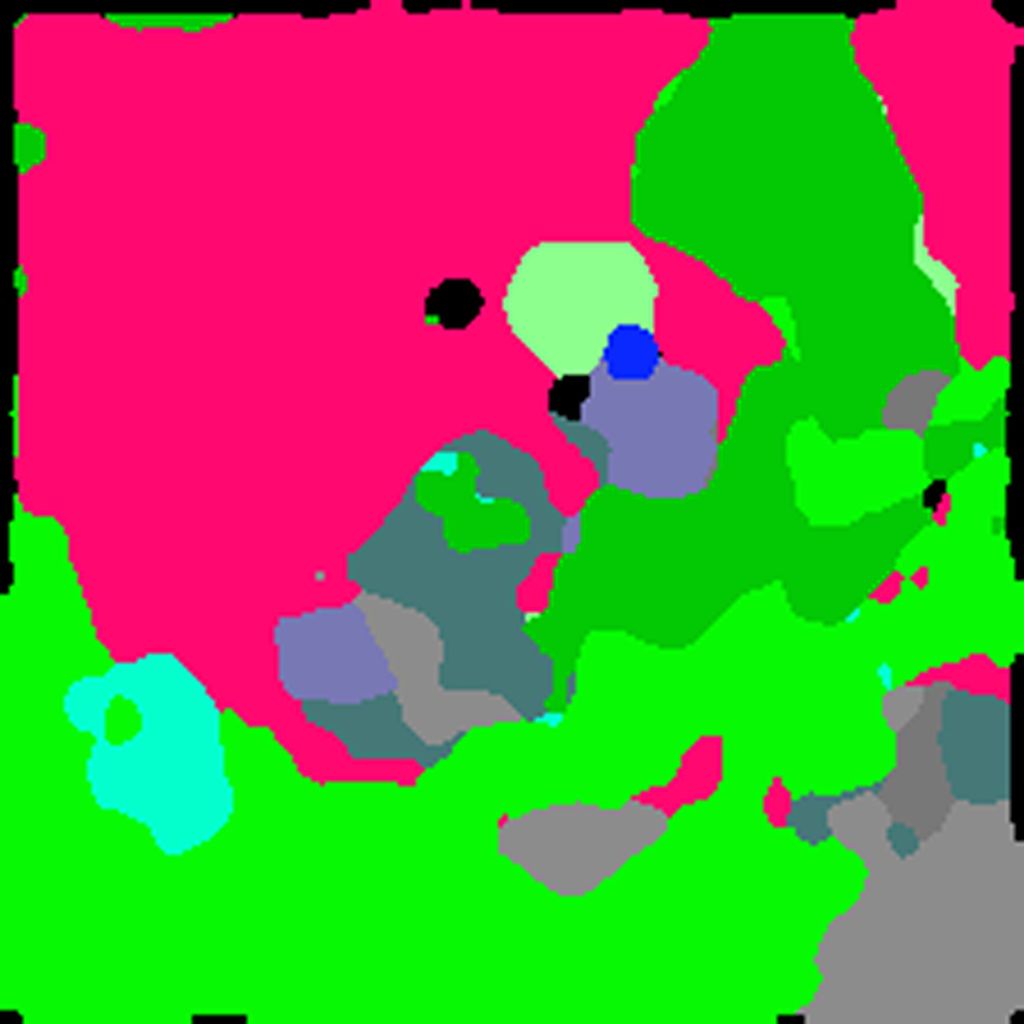}  &
      \includegraphics[height=5cm,keepaspectratio]{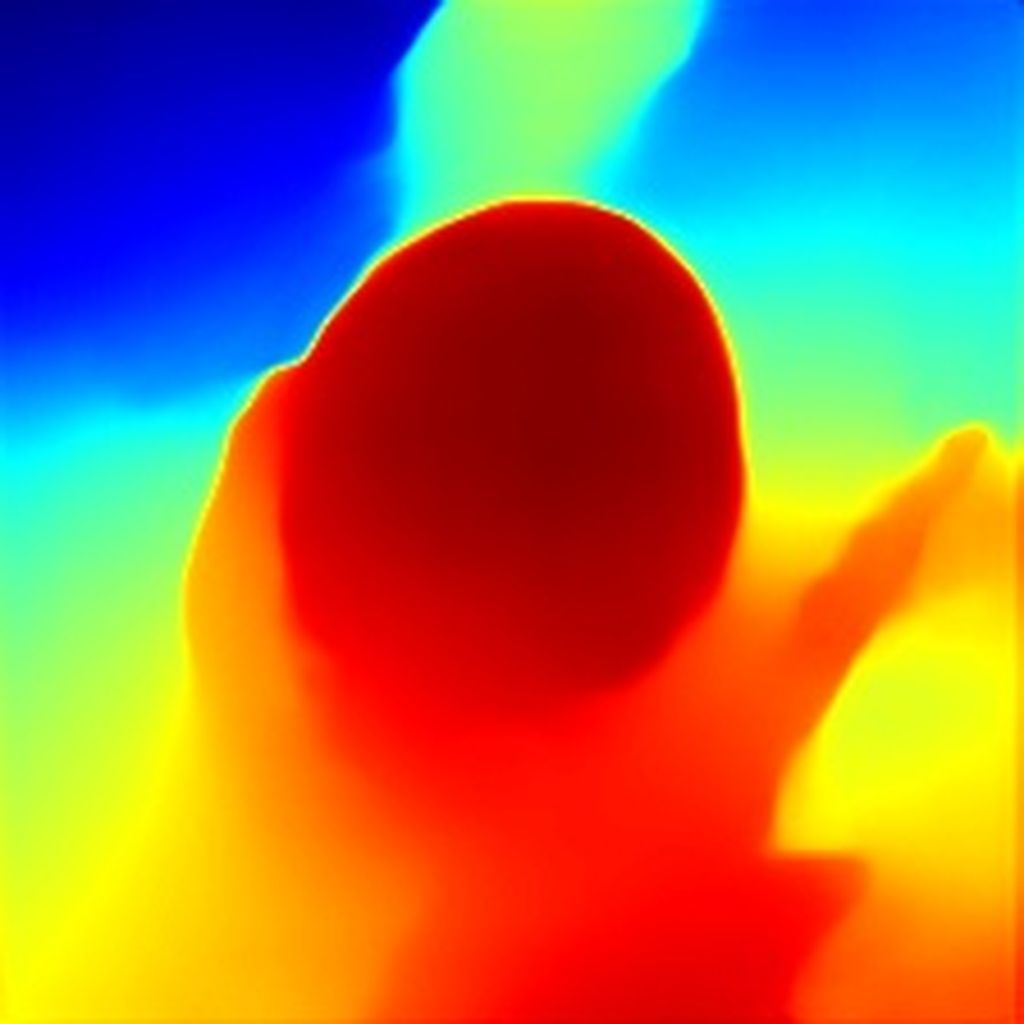} & 
      \includegraphics[height=5cm,keepaspectratio]{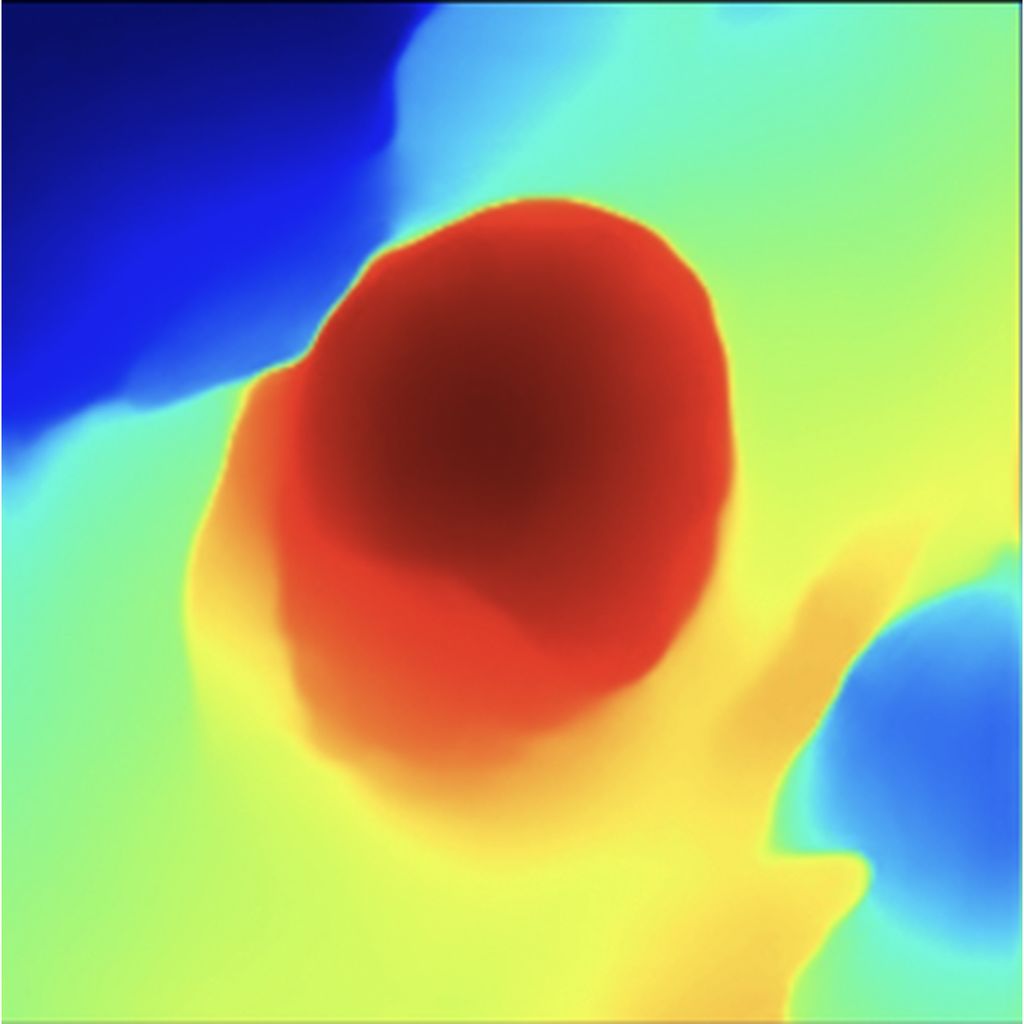} & 
      \includegraphics[height=5cm,keepaspectratio]{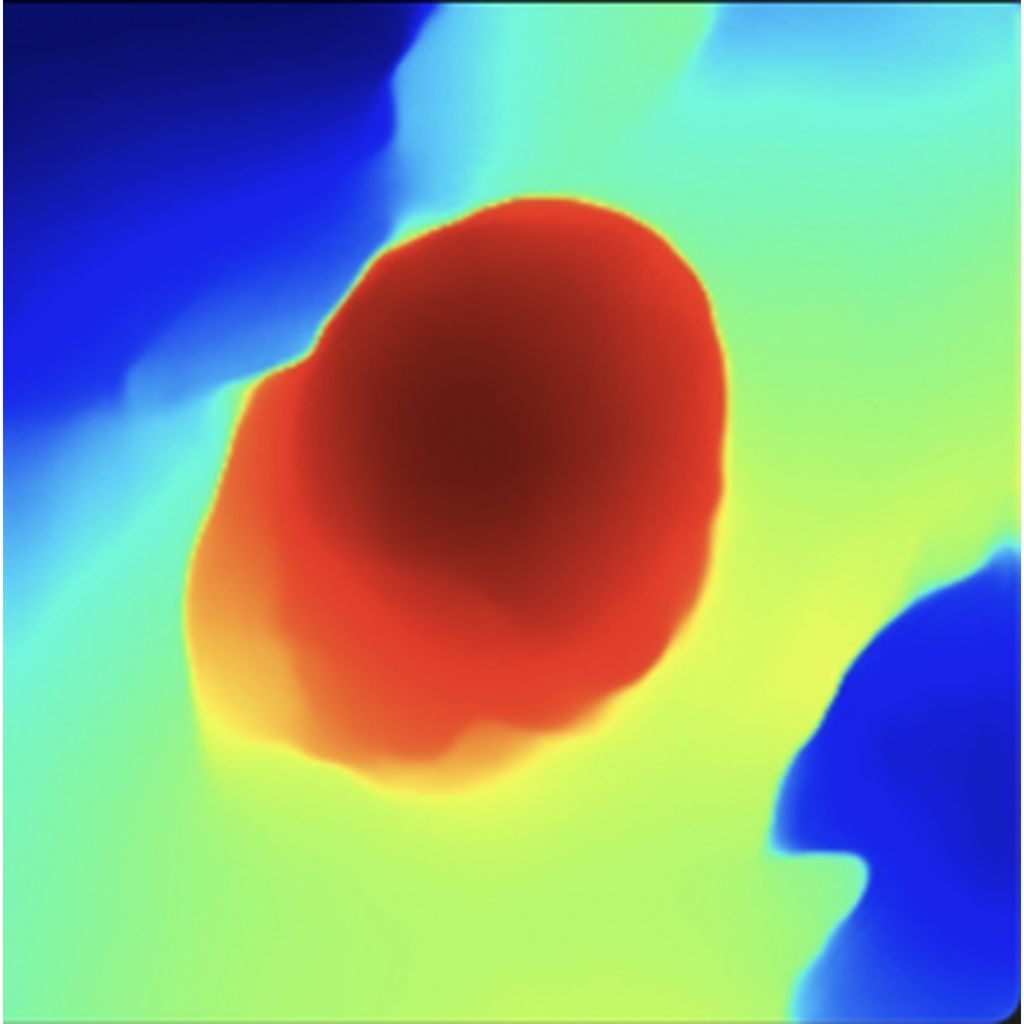} \\
    \end{tabular}%
  }

    \caption{Effectiveness of invariant-based regularization in improving image colorization. {Left panel}: Comparison using segmentation invariant. {Right panel}: Comparison using depth invariant.
Each panel shows the colorized outputs (top row) and corresponding invariant masks (bottom row). Invariant-regularized models significantly preserve critical structural details, such as object boundaries and spatial depth cues, compared to baseline models that exhibit noticeable artifacts and inaccuracies.}
  
  \label{fig:color_comparison}
  
\end{figure*}

The aim of training with invariants is to encourage the trained model to produce predictions that, while they may not perfectly match the ground truth, deviate from the ground truth in ways that are approximately preserved by the invariant.  The question that we seek to answer in this section, then, is to what extent invariance preservation is reflected in existing metrics.  In particular, we are interested in understanding (1) to what extent each metric is sensitive to differences with respect to the selected invariants and (2) to what degree enforcing invariants yields performance improvements in practice.

\begin{table*}[t!]
\centering
\scriptsize
\setlength{\tabcolsep}{2pt}

\begin{tabular}{ m{8em} | m{20em} |  m{20em} |  m{20em} }
\toprule
\textbf{Task} & Image Colorization & German to English Neural Machine Translation & Image Captioning\\
\hline
\textbf{Base Networks}  &  ChromaGAN \citep{vitoria2020chromagan}, ColorFormer \citep{ji2022colorformer}, and DDColor \citep{kang2023ddcolor} & BiBERT \citep{xu2021bert} and Bi-SumCut \citep{gao2022bi} & SmallCap \citep{ramos2023smallcap} and ClipCap \citep{mokady2021clipcap}\\
\hline
\textbf{Loss Function ($L_T$)} & Pixel‐reconstruction MSE, L1-loss, Perceptual-loss, GAN-loss, Colorfulness-loss  & Label Smoothed Cross Entropy: Cross-entropy loss measuring predicted vs target word probability differences with label smoothing for better generalization
 & Cross Entropy Loss: Token-level classification loss between predicted and ground truth caption sequences\\
\hline
\textbf{Invariant Networks} & \multirow{2}{23em}{Depth: MiDaS \citep{Ranftl2020}\\ Segmentation: Side Adapter Network (SAN)\\\citep{xu2023side}} & \multirow{2}{23em}{NER:  Hugging Face's \citep{wolf2019huggingface} \\ Sentiment: Hugging Face's distilbert-base-\\uncased-finetuned-sst-2-english} & \multirow{2}{24em}{NER:  Hugging Face's \citep{wolf2019huggingface} \\ Scenegraph: sngparser \citep{wu2019unified}\\\citep{schuster2015generating}}\\ & & & \\ & & & \\ %
\hline
\textbf{Invariant Loss ($L_I$)} & SegmentationSANLoss: Dice loss between predicted vs GT SAN segmentation masks prevents color bleeding
DepthMidasLoss: SILog loss between predicted vs GT MiDaS depth maps preserves spatial structure  & NER Loss: KL divergence between predicted vs GT BERT-NER token classifications for entity preservation and Sentiment Loss: KL divergence between predicted vs GT DistilBERT sentiment logits & NER Loss: KL divergence between predicted vs GT BERT-NER token classifications for entity preservation.   
Scene Graph Loss: Symmetric difference ratio of entities and relations between predicted vs GT scene graphs\\
\hline
\textbf{Metrics} &  Frechet Inception Distance (FID), Frechet DinoV2 Distance (FDD), CLIP maximum mean discrepancy distance (CMMD), Learned Perceptual Image Patch Similarity (LPIPS), Colorfulness (CF), Peak Signal to Noise Ratio (PSNR)  & FBD-4(fbd of average of last 4 hidden layers), Frechet Bert Distance(FBD) \citep{xiang2021assessing}, BERTScore \citep{zhang2020bertscore}, BLEU-4\citep{papineni2002bleu}, METEOR \citep{banerjee2005meteor},  COMET \citep{rei2020comet}  & FBD-4(fbd of average of last 4 hidden layers),  FBD \citep{xiang2021assessing}, BERTScore \citep{zhang2020bertscore}, CIDEr \citep{vedantam2015cider}, BLEU-4\citep{papineni2002bleu}, METEOR  \citep{banerjee2005meteor}\\
\hline
\textbf{Data Sets} & ImageNet \citep{deng2009imagenet}, COCO \citep{lin2014microsoft}, and ADE20K \citep{zhou2019semantic} \citep{zhou2017scene} & IWSLT 2014 \citep{cettolo2012wit3} and WMT-14 \citep{luong2015stanford}& COCO \citep{lin2014microsoft} and flickr30k \citep{plummer2015flickr30k}\\
\midrule
\bottomrule
\end{tabular}
\caption{Experimental configurations for the three selected tasks.}
\label{table:taskconfig}
\end{table*}

 We illustrate our framework by applying it to the image colorization, machine translation, and image captioning tasks. All models were trained from scratch to ensure that the same training process was used for both the baseline models and the baseline+invariant models.  Note that in all experimental settings the weights of the invariant networks are frozen during training. Full implementation details for the baselines can be found in the original papers, and all architectures were obtained directly from the authors of these works. All systems were trained using either one NVIDIA A100 GPU or one V100 GPU. In Table \ref{table:taskconfig}, we summarize the choice of baselines, invariants, data sets, losses, and metrics for each experimental setting. 
 Full experimental details can be found in the appendix.

\begin{figure*}[t!]
 \centering
 \setlength{\tabcolsep}{0pt}
 \renewcommand{\arraystretch}{0}

\resizebox{\textwidth}{!}{
    \begin{tabular}{ccc}        
     \includegraphics{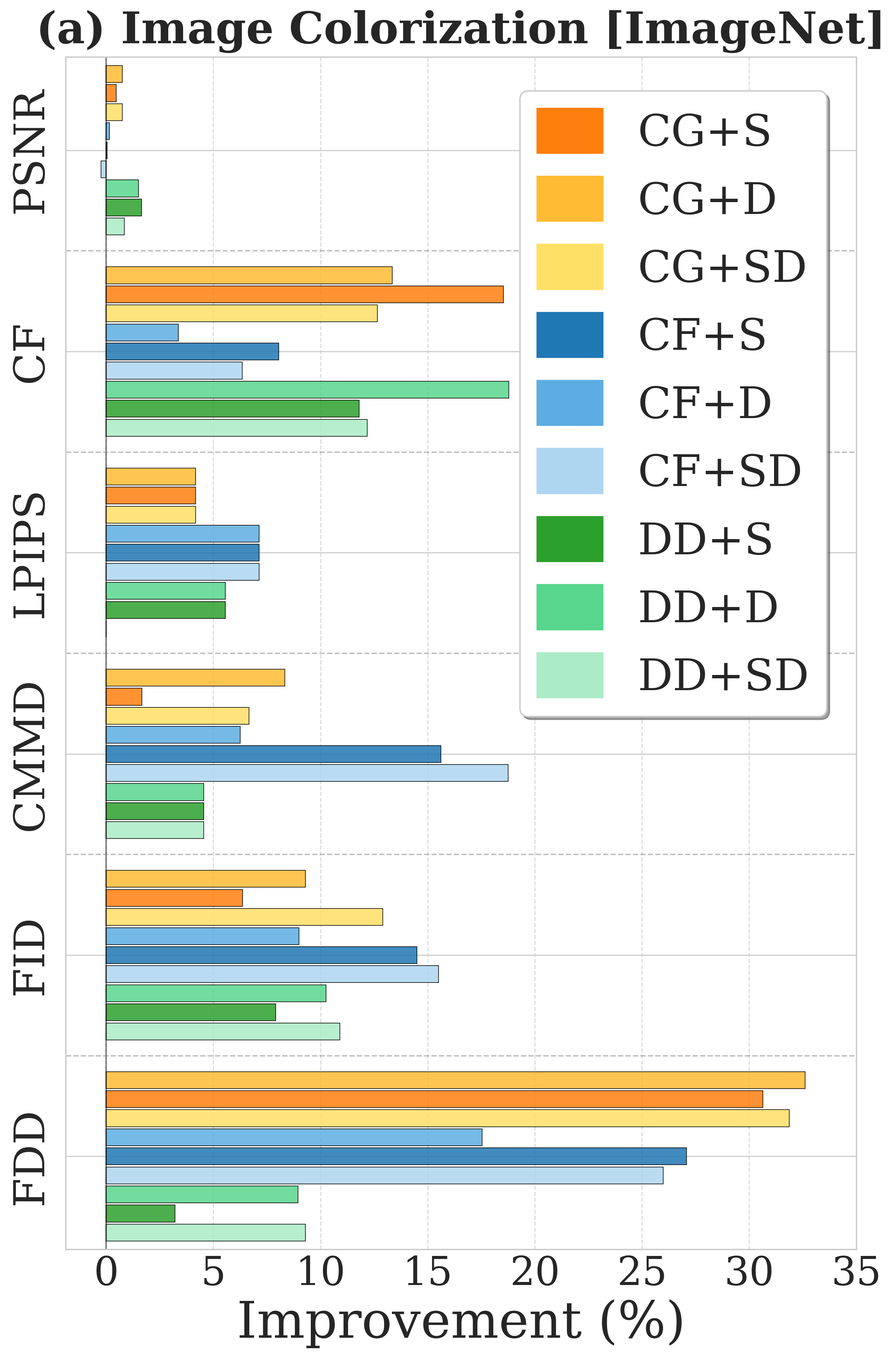}  &
     \includegraphics{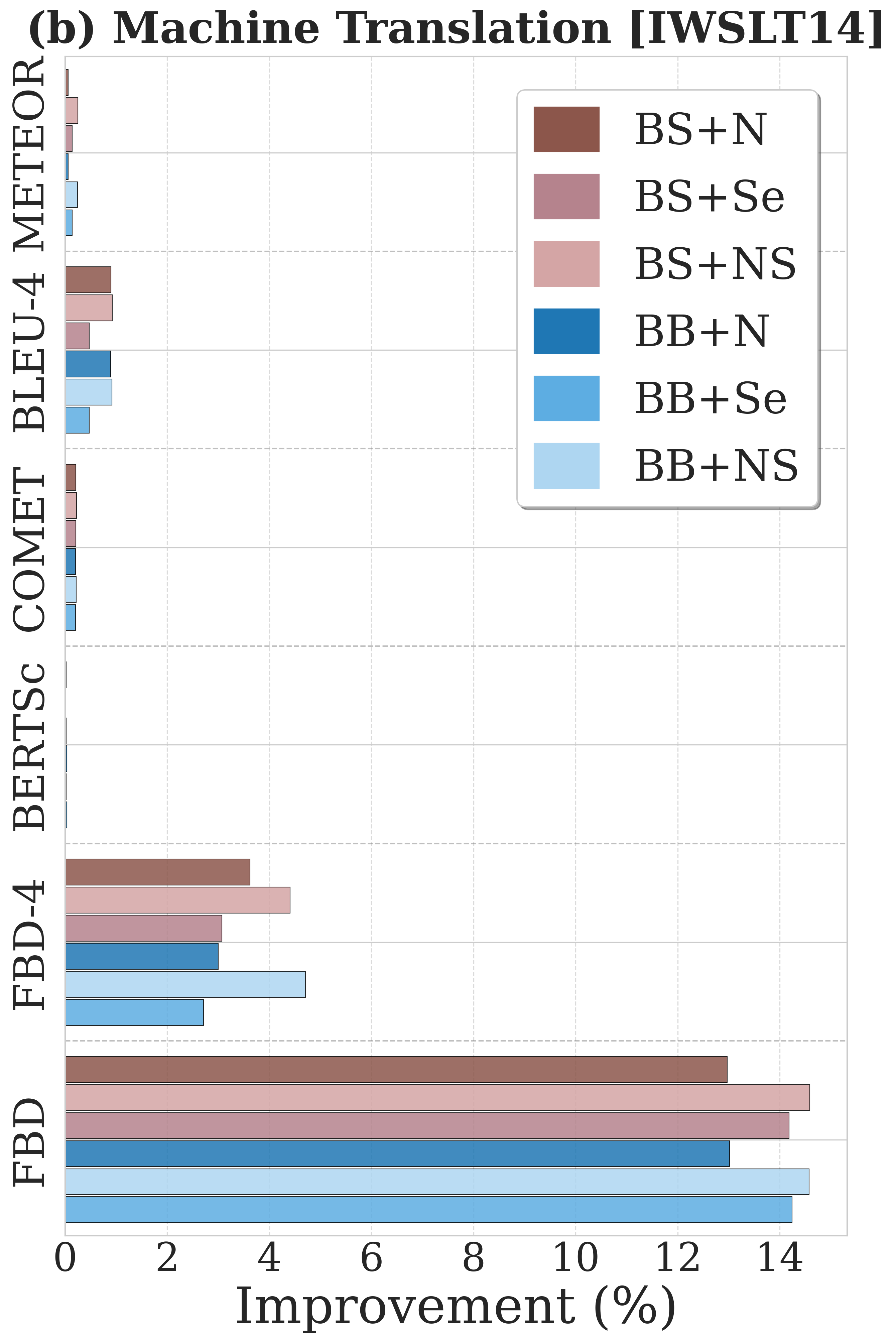}  &
     \includegraphics{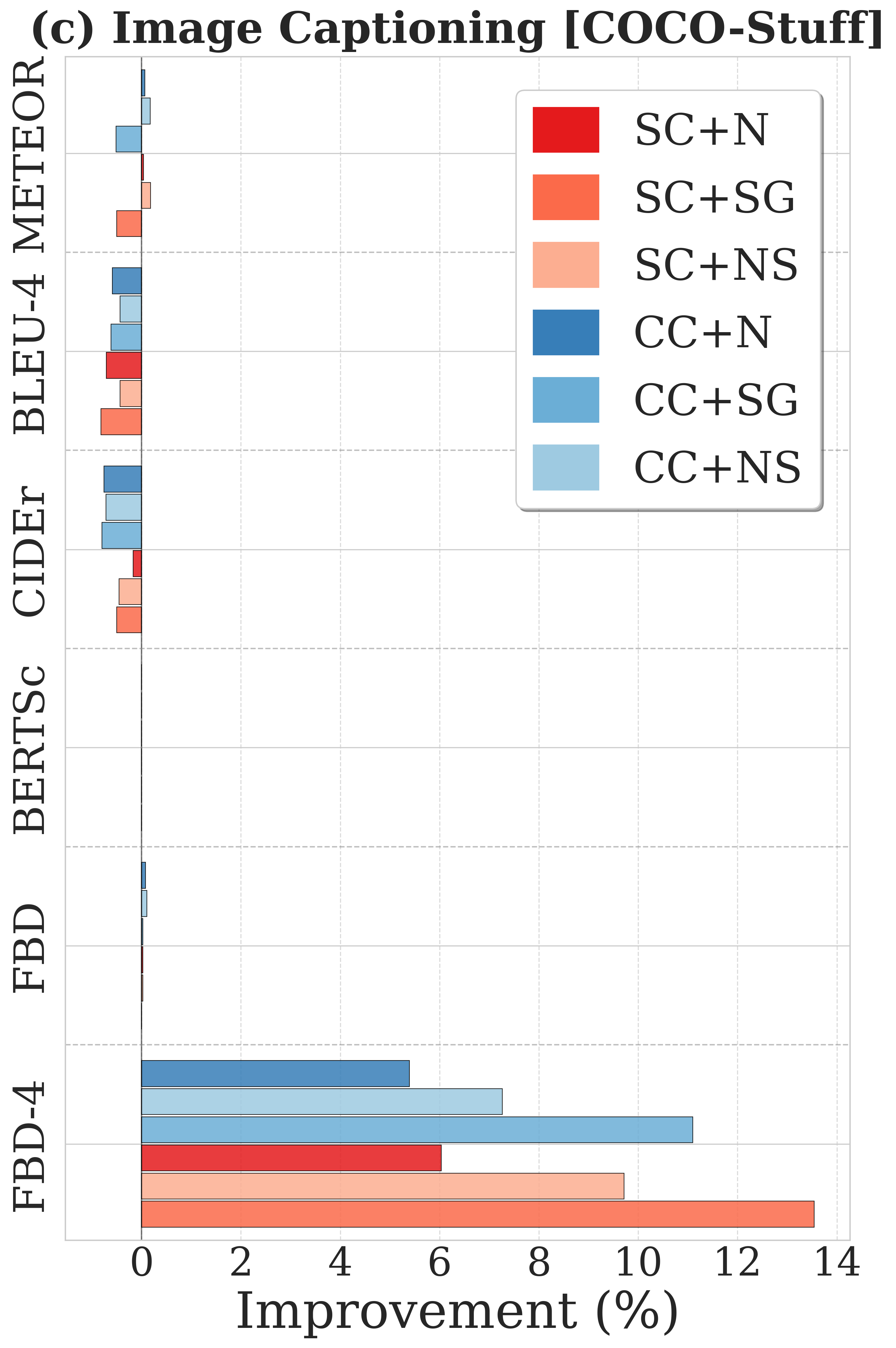}
     \\  
   \end{tabular}
}

 \caption{Performance improvement of architectures trained with invariants versus training without invariants, evaluated across common metrics for three tasks. 
(a) Image Colorization: Base models are ChromaGAN (CG), ColorFormer (CF), and DDColor (DD), each augmented with Segmentation (+S), Depth (+D), or both (+SD). 
(b) Machine Translation: Base models are Bi-SimCut (BS) and BiBERT (BB), augmented with NER (+N), SeAM (+Se), or both (+NS). 
(c) Image Captioning: Base models are SmallCap (SC) and ClipCap (CC), augmented with NER (+N), SceneGraph (+SG), or both (+NS). 
Distributional metrics (FDD, FBD, FID, CMMD) show larger improvements than traditional metrics (PSNR, BLEU-4, METEOR), indicating higher sensitivity to invariant-based training.}

 \label{fig:barcharts}
\end{figure*}

\begin{table*}
\centering

\scriptsize
\setlength{\tabcolsep}{2.5pt}
\begin{tabular}{l c | c c c c c c | c c c c c c}
\toprule
\multirow{2}{*}{Method} & \multirow{2}{*}{\#P} & \multicolumn{6}{c|}{ImageNet (val50k)} & \multicolumn{6}{c}{COCO-Stuff(val5k)} \\
\cmidrule(lr){3-8} \cmidrule(lr){9-14}
& & \textbf{FDD}$\downarrow$ & FID$\downarrow$ & CMMD$\downarrow$ & LPIPS$\downarrow$ & CF$\uparrow$ & PSNR$\uparrow$ & FDD$\downarrow$ & FID$\downarrow$ & CMMD$\downarrow$ & LPIPS$\downarrow$ & CF$\uparrow$ & PSNR$\uparrow$ \\
\midrule
ChromaGAN & 173.9M & \textbf{42.36} & 5.81 & 0.597 & 0.236 & 27.41 & 23.65 & 94.90 & 16.73 & 0.715 & 0.233 & 27.89 & 23.66 \\
\hspace{0.6em}+Seg & 173.9M & \textbf{29.38} & 5.44 & 0.679 & 0.236 & 29.36 & 23.76 & 81.07 & 15.50 & 0.603 & 0.233 & 29.56 & 23.82 \\
\hspace{0.6em}+Depth & 173.9M & \textbf{28.55} & 5.27 & 0.565 & 0.231 & 28.73 & 23.83 & 79.28 & 15.29 & 0.659 & 0.229 & 28.95 & 23.91 \\
\hspace{0.6em}+Seg+Depth & 173.9M & \textbf{28.86} & 5.06 & 0.599 & 0.231 & 28.69 & 23.83 & 80.00 & 14.58 & 0.597 & 0.229 & 29.99 & 23.82 \\
\midrule
ColorFormer & 44.8M & \textbf{21.16} & 2.00 & 0.318 & 0.137 & 27.57 & 24.91 & 59.38 & 9.03 & 0.377 & 0.134 & 28.30 & 24.70 \\
\hspace{0.6em}+Seg & 44.8M & \textbf{15.43} & 1.71 & 0.273 & 0.130 & 28.29 & 24.92 & 52.52 & 8.49 & 0.370 & 0.134 & 29.21 & 24.67 \\
\hspace{0.6em}+Depth & 44.8M & \textbf{17.45} & 1.82 & 0.302 & 0.129 & 28.16 & 24.95 & 55.31 & 8.80 & 0.385 & 0.133 & 28.83 & 24.73 \\
\hspace{0.6em}+Seg+Depth & 44.8M & \textbf{15.66} & 1.69 & 0.263 & 0.131 & 28.52 & 24.85 & 52.74 & 8.45 & 0.376 & 0.135 & 29.39 & 24.61 \\
\midrule
DDColor & 55M & \textbf{29.60} & 4.68 & 0.436 & 0.179 & 52.84 & 21.22 & 71.58 & 12.08 & 0.484 & 0.177 & 50.88 & 21.43 \\
\hspace{0.6em}+Seg & 55M & \textbf{28.65} & 4.31 & 0.423 & 0.174 & 51.14 & 21.57 & 69.70 & 11.48 & 0.465 & 0.172 & 50.10 & 21.73 \\
\hspace{0.6em}+Depth & 55M & \textbf{26.95} & 4.20 & 0.418 & 0.174 & 51.63 & 21.54 & 69.69 & 11.22 & 0.462 & 0.172 & 50.27 & 21.73 \\
\hspace{0.6em}+Seg+Depth & 55M & \textbf{26.85} & 4.17 & 0.419 & 0.175 & 52.12 & 21.40 & 69.35 &11.20  & 0.465 & 0.172 & 50.67 & 21.63 \\
\bottomrule
\end{tabular}
\caption{Quantitative comparison of image colorization methods. $\uparrow$ ($\downarrow$) indicates higher (lower) values are better; {Bold} values in FDD column of ImageNet indicate the model selection criterion. The Seg architecture is SAN, and depth architecture is Midas.} 
\label{table:colorcomparison}
\end{table*}

\begin{table*}[t]
\centering
\scriptsize
\setlength{\tabcolsep}{2pt}
\begin{tabular}{@{}l c | c c c c c c | c c c c c c@{}}
\toprule
\multirow{2}{*}{Model} & \multirow{2}{*}{\#P} & \multicolumn{6}{c|}{COCO-Stuff (val5k)} & \multicolumn{6}{c}{Flickr30k (val1k)} \\
\cmidrule(lr){3-8} \cmidrule(lr){9-14}
& & \textbf{FBD-4$\downarrow$} & FBD$\downarrow$ & BERTScore$\uparrow$ & CIDEr$\uparrow$ & BLEU-4$\uparrow$ & METEOR$\uparrow$ & FBD-4$\downarrow$ & FBD$\downarrow$ & BERTScore$\uparrow$ & CIDEr$\uparrow$ & BLEU-4$\uparrow$ & METEOR$\uparrow$ \\
\midrule
SmallCap & 7M & \textbf{62.60} & 41.90 & 0.9224 & 118.49 & 36.68 & 27.90 & 58.45 & 39.12 & 0.9189 & 112.30 & 34.82 & 26.45 \\
\hspace{1em}+NER & 7M & \textbf{58.82} & 41.89 & 0.9223 & 116.29 & 36.07 & 27.91 & 55.20 & 39.08 & 0.9187 & 110.15 & 34.25 & 26.48 \\
\hspace{1em}+SceneGraph & 7M & \textbf{54.12} & 41.90 & 0.9218 & 115.47 & 35.68 & 27.76 & 51.85 & 39.15 & 0.9182 & 109.42 & 33.89 & 26.32 \\
\hspace{1em}+NER+SceneGraph & 7M & \textbf{56.52} & 41.89 & 0.9226 & 117.48 & 36.52 & 27.95 & 53.78 & 39.11 & 0.9191 & 111.68 & 34.71 & 26.51 \\
\midrule
ClipCap & 43M & \textbf{48.20} & 38.45 & 0.9301 & 125.80 & 39.15 & 29.20 & 44.60 & 36.25 & 0.9267 & 119.45 & 37.28 & 27.85 \\
\hspace{1em}+NER & 43M & \textbf{45.60} & 38.42 & 0.9299 & 123.45 & 38.52 & 29.22 & 42.15 & 36.21 & 0.9265 & 117.30 & 36.65 & 27.88 \\
\hspace{1em}+SceneGraph & 43M & \textbf{42.85} & 38.44 & 0.9295 & 122.30 & 38.01 & 29.05 & 39.85 & 36.28 & 0.9260 & 116.52 & 36.14 & 27.72 \\
\hspace{1em}+NER+SceneGraph & 43M & \textbf{44.70} & 38.41 & 0.9303 & 124.90 & 38.98 & 29.25 & 41.25 & 36.24 & 0.9269 & 118.85 & 37.12 & 27.91 \\
\bottomrule
\end{tabular}
\caption{Performance on image captioning benchmarks.  {Bold} values in FBD-4 column of COCO-Stuff indicate the model selection criterion. }

\label{table:captioning}
\end{table*}

\begin{table*}[t]
\centering
\scriptsize
\setlength{\tabcolsep}{2pt}

\begin{tabular}{@{}l c | c c c c c c | c c c c c c@{}}
\toprule
\multirow{2}{*}{Model} & \multirow{2}{*}{\#P} & \multicolumn{6}{c|}{IWSLT-14} & \multicolumn{6}{c}{WMT-14} \\
\cmidrule(lr){3-8} \cmidrule(lr){9-14}
& & \textbf{FBD$\downarrow$} & FBD-4$\downarrow$ & BERTScore$\uparrow$ & BLEU-4$\uparrow$ & METEOR$\uparrow$ & COMET$\uparrow$ & FBD$\downarrow$ & FBD-4$\downarrow$ & BERTScore$\uparrow$ & BLEU-4$\uparrow$ & METEOR$\uparrow$ & COMET$\uparrow$\\
\midrule
Bi-SimCut & 65M & \textbf{9.87} & 102.15 & 0.8871 & 37.83 & 0.6412 & 0.5743 & 15.24  & 135.42 & 0.8952 & 34.47 & 0.5836 & 0.6124 \\
\hspace{1em}+NER         & 65M  & \textbf{8.42} & 101.89 & 0.8867 & 36.95 & 0.6358 & 0.5739 & 14.18 & 134.76 & 0.8948 & 33.82 & 0.5794 & 0.6119 \\
\hspace{1em}+SeAM        & 65M  & \textbf{8.51} & 103.14 & 0.8868 & 37.02 & 0.6361 & 0.5737 & 14.26 & 136.23 & 0.8949 & 33.89 & 0.5798 & 0.6117 \\
\hspace{1em}+NER+SeAM    & 65M  & \textbf{8.38} & 102.67 & 0.8869 & 37.18 & 0.6373 & 0.5741 & 14.09 & 135.58 & 0.8950 & 34.06 & 0.5809 & 0.6121 \\
\midrule
BiBERT & 110M & \textbf{8.99} & 100.24 & 0.8930 & 38.20 & 0.6597 & 0.5899 & 13.76 & 133.18  & 0.9012 & 35.24 & 0.6023 & 0.6287 \\
\hspace{1em}+NER         & 110M & \textbf{7.71} & 100.03 & 0.8925 & 37.35 & 0.6538 & 0.5894 & 12.84 & 132.91 & 0.9008 & 34.52 & 0.5976 & 0.6282 \\
\hspace{1em}+SeAM        & 110M & \textbf{7.82} & 101.27 & 0.8926 & 37.41 & 0.6536 & 0.5890 & 12.93 & 134.17 & 0.9009 & 34.58 & 0.5974 & 0.6279 \\
\hspace{1em}+NER+SeAM    & 110M & \textbf{7.68} & 100.80 & 0.8927 & 37.58 & 0.6548 & 0.5897 & 12.76 & 133.75 & 0.9010 & 34.74 & 0.5987 & 0.6284 \\
\bottomrule
\end{tabular}
\caption{Performance on machine translation. {Bold} values indicate the model selection criterion.}
\label{table:nlpcomparison_appendix}
\end{table*}

\subsection{Sensitivity of Metrics to Invariants}

Ideally, we would want a model for the selected tasks that produces labels that align well with the ground truth, both in terms of the loss used for training and the chosen invariant loss.  However, different evaluation metrics may be more or less sensitive to changes with respect the invariant loss while the training loss remains stable.  This could happen for a variety of reasons.  For example, consider the FID metric that has gained popularity as an evaluation metric for colorization models.  The loss fits a multivariate Gaussian distribution over a feature space extracted from the Inception network.  As this feature space is much lower dimensional than the input images, some information is inherently lost.  For example, it could be that two colorized images with the same MSE to the ground truth, but different invariant losses, map to the same Inception features.   If this occurs, the differences being captured by the invariant loss have been forgotten in the feature space used for evaluation.  In this instance, we would say that the Inception features are not \textit{sensitive} to the selected invariant.  

As our primary supposition in this work is that invariants are properties that should be approximately preserved by ``good'' labels, it is critical that the evaluation metrics be sensitive to the selected invariant(s).  Our first set of experiments is designed to assess the sensitivity of a variety of evaluation metrics to a selected set of invariants for each of our three chosen tasks.  To this end, we trained a variety of architectures both with and without an invariant loss and evaluated them using a variety of metrics on validation data.  

Figure \ref{fig:barcharts} illustrates the best possible performance improvement (compared against the baseline with no invariant loss during training) for each invariant/metric pair obtained via hyperparameter search on the indicated validation data sets for our chosen tasks.  Note that only $\alpha$, the hyperparameter for the invariant loss was tuned here, all other settings for the architectures remained untouched from the baseline. For the image captioning task, even a small invariant penalty during training results in $\pm 1\%$ performance improvement / degradation for METEOR, BLEU-4, and CIDEr.  FBD and BertScore show essentially no differences with respect to training with/without an invariant penalty.  This suggests that these metrics are not very sensitive to the selected invariants for this task.  However, FBD-4, constructed from the last 4 layers of the BERT network, shows $5\%-13\%$ improvement across models/invariants.   For machine translation, METEOR, BLEU-4, CIDEr, and BERTScore all show a $<1\%$ improvement while FBD-4 improves $2\%-5\%$ and FBD improves $13\%-15\%$. A similar trend can be observed for the colorization task:  PSNR, a traditional metric, shows a $<2\%$ improvement, LPIPS shows moderate $4\%-7\%$ improvement, FID shows $7\%-16\%$ improvement, and FDD shows a more dramatic $3\%-33\%$ improvement. In all cases, metrics derived from richer feature embeddings tend to show increased sensitivity to training with invariants.

From these observations, we can conclude that, while training with an invariant penalty does clearly result in different predictions, many of the standard error metrics for evaluating task performance are almost insensitive to these changes compared with distributional metrics and metrics derived from richer feature spaces. Put another way, if we had valdiated a model using the insensitive metrics, we essentially would not be able to tell whether or not the selected model respects the invariant.  This suggests that the natural invariants considered here are not being automatically enforced through the standard train/validate pipeline.

\subsection{The Effect of Invariants on Task Performance}

\begin{figure*}[ht!]
  \centering

  \begin{minipage}[t]{0.36\textwidth}
    \raisebox{-0.5\height}{\includegraphics[clip,width=\linewidth]{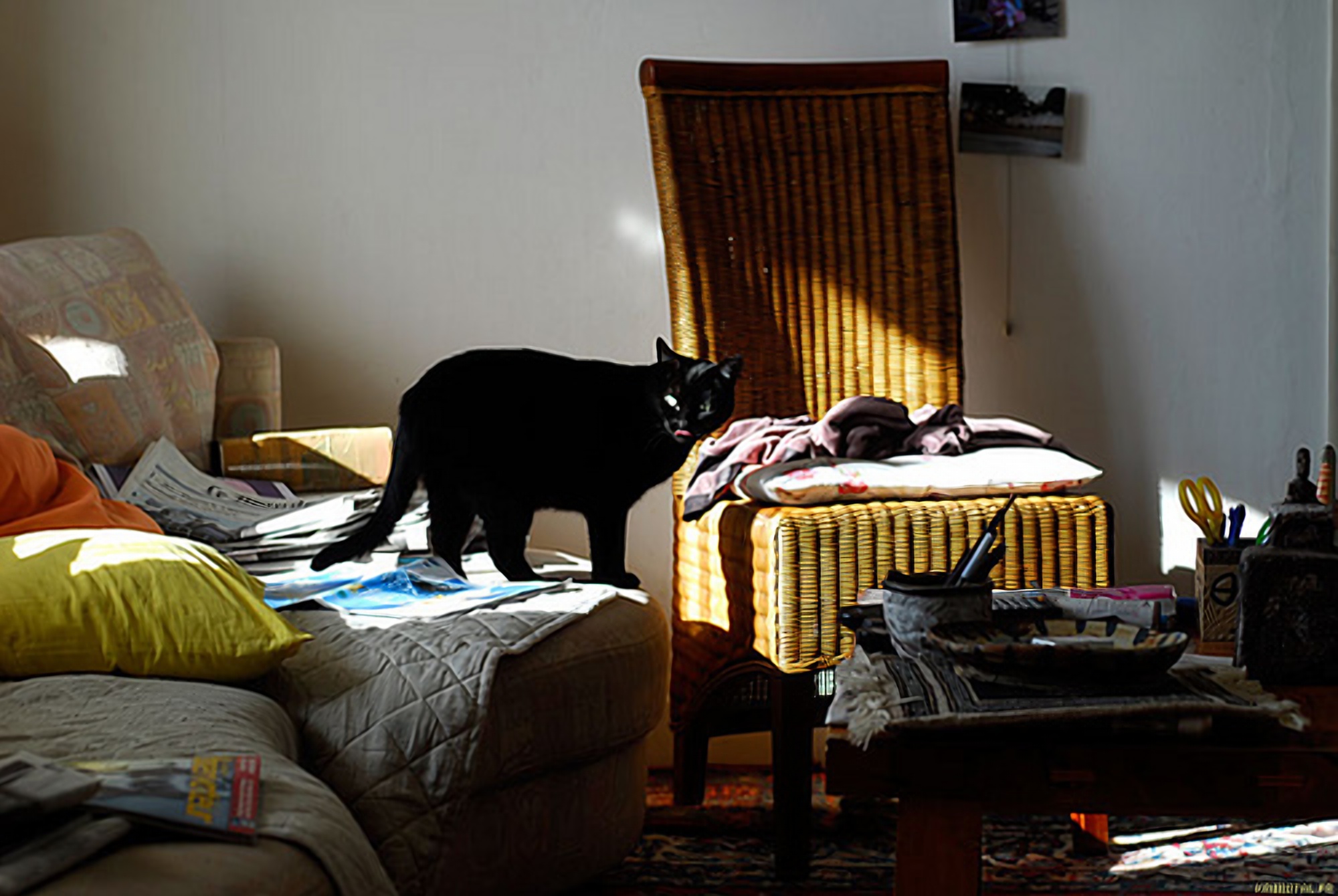}}
  \end{minipage}%
  \hspace{.1em}
  \begin{minipage}[t]{0.62\textwidth}
  \centering
    \scriptsize
    \setlength{\tabcolsep}{1.5pt}
    \renewcommand{\arraystretch}{0.9}
    \begin{tabular}{|p{0.30\textwidth}|p{0.30\textwidth}|p{0.30\textwidth}|}
    \hline
    \textbf{Ground Truth} & \textbf{SmallCap+NER+SceneGraph} & \textbf{SmallCap} \\ \hline
    
    \itshape Caption: A cat standing on a couch in a cluttered living room. &
    \itshape Caption: A black cat standing on a couch in a living room. &
    \itshape Caption: A black cat standing on top of a bed. \\ \hline
    
    \textbf{Entities:} & \textbf{Entities:} & \textbf{Entities:} \\
    \begin{tabular}[t]{@{}p{0.08\textwidth}p{0.11\textwidth}p{0.08\textwidth}@{}}
    Head & Span & Modifiers \\ \hline
    cat & a cat & a \\
    couch & a couch & a \\
    living room & a cluttered living room & a, cluttered \\
    \end{tabular} &
    \begin{tabular}[t]{@{}p{0.08\textwidth}p{0.11\textwidth}p{0.08\textwidth}@{}}
    Head & Span & Modifiers \\ \hline
    cat & a black cat & a, black \\
    couch & a couch & a \\
    living room & a living room & a \\
    \end{tabular} &
    \begin{tabular}[t]{@{}p{0.08\textwidth}p{0.11\textwidth}p{0.08\textwidth}@{}}
    Head & Span & Modifiers \\ \hline
    cat & a black cat & a, black \\
    bed & a bed & a \\
    \vspace*{-\baselineskip}\mbox{} & & \\
    \end{tabular} \\ \hline
    
    \textbf{Relations:} & \textbf{Relations:} & \textbf{Relations:} \\
    \begin{tabular}[t]{@{}p{0.09\textwidth}p{0.09\textwidth}p{0.09\textwidth}@{}}
    Subject & Relation & Object \\ \hline
    cat & on & couch \\
    cat & in & living room \\
    \end{tabular} &
    \begin{tabular}[t]{@{}p{0.09\textwidth}p{0.09\textwidth}p{0.09\textwidth}@{}}
    Subject & Relation & Object \\ \hline
    cat & on & couch \\
    cat & in & living room \\
    \end{tabular} &
    \begin{tabular}[t]{@{}p{0.09\textwidth}p{0.09\textwidth}p{0.09\textwidth}@{}}
    Subject & Relation & Object \\ \hline
    cat & on & bed \\
    \vspace*{-\baselineskip}\mbox{} & & \\
    \end{tabular} \\ \hline
    \end{tabular}
  \end{minipage}
  \caption{Image captions produced by models trained with/without scenegraph and NER invariants for an image of a cat (left).}
  \label{fig:caption_comparison}
\end{figure*}

Above, we observed that distributional metrics seem to be more sensitive to training with invariants than their traditional counterparts when we consider the best performance that can be obtained on validation data.  This sensitivity is important if we care about preserving the invariants in practice, but high sensitivity alone does not guarantee that a metric is necessarily ``good'' for evaluating task performance.  In particular, models that perform the best along one metric could perform poorly on all other metrics.  In this case, high performance was achieved along one metric at the expense of the other metrics.  In this section, we explore whether or not training with an invariant penalty yields performance improvements more broadly.  That is, does incorporating task invariants as above and validating with an invariant-sensitive metric yield practical performance gains? 

For each of our tasks, we trained a variety of architectures with different invariant losses and report the resulting performance across common task metrics.  The best model for each architecture/invariant combination was selected using the most sensitive metric for each task from the previous section, i.e., FDD for colorization, FBD-4 for image captioning, and FBD for machine translation, on a validation data set.   The results of our investigation can be found in Tables \ref{table:colorcomparison} and \ref{table:captioning}. %
As was noted above, the best performing models each perform better with respect to the metric used for model selection.  However, the more detailed results in the tables illustrate that the performance improvements are not accompanied by significant degredation in the other metrics.  For example, in Table \ref{table:colorcomparison}, Colorformer+Seg achieves the best FDD score across all architectures and yields performance improvements on nearly all metrics across the two validation data sets.  Contrast this with the results in Table \ref{table:captioning} in which dramatic improvements in FBD-4 are met with minor $\pm2\%$ differences in nearly all other metrics.  A similar observation can be made for FBD in machine translation.

The tables also illustrate the effect of using more than one invariant penalty during training. Recall that, as the invariant networks are frozen during training, the only additional training overhead comes from evaluating these networks.  In colorization, ColorFormer+Seg+Depth acheives a lower FDD than ColorFormer+Depth, but is slightly worse than ColorFormer+Seg. A similar observation can be made for ClipClap+NER+Scenegraph for image captioning.  In machine translation, BiBERT+NER+SeAM achieves the best overall FBD score.  Overall, while adding two potentially conflicting losses can lead to optimization challenges, the results still remain competitive while encouraging preservation of more than one invariant.

Qualitatively, we have found that training with invariants can help correct errors that violate the invariants.  In Figure \ref{fig:color_comparison}, we provide an example of colorization outputs when trained with/without a depth or a segmentation invariant.  In both cases we observe that better aligned outputs from the invariant networks correlate with better colorizations.  In Figure \ref{fig:caption_comparison}, an example of image captioning with an NER+SceneGraph invariant shows that a better caption is produced and the SceneGraph generated from the ground truth image better matches the SceneGraph from the NER+SceneGraph invariant model (the invaraint-trained model correctly identifies the couch).   

Additional qualitative results for all of the tasks can be found in the appendix.

 \section{Conclusion}
In summary, our experimental results illustrate that (1) traditional metrics, e.g., PSNR, BLEU-4, etc., show minimal sensitivity to invariants, (2) distributional metrics, e.g., FID, FDD, and FBD variants, show substantial improvements when invariants are enforced without significant loss with respect to traditional metrics, and (3) richer feature representations, e.g., DinoV2 in FDD, show even greater sensitivity to invariants than less rich representations.  Critically, our metric evaluation framework does not depend on human evaluators, which makes this a flexible approach that extends beyond the tasks considered here. 
Furthermore, our results suggest that future metric design should favor richer feature encoders, e.g., DinoV2 rather than Inception for images, and consider incorporating explicit invariant-violation penalties.

\section{Limitations and Future Work}
Our framework relies on approximate task invariants derived from pretrained models, which may introduce biases or fail in low-resource domains where such models are unavailable. Moreover, we do not directly validate the hypothesis that enforcing approximate invariants leads to improved human evaluation; instead, we position invariants as a scalable diagnostic when human studies are impractical. Finally, our empirical study focuses on three domains (image colorization, machine translation, and image captioning). Extending invariants-as-probes to other multi-output structured tasks and integrating them directly into metric design are interesting directions for future work.

\section*{Acknowledgements}
This work was supported in part by NSF grant 2327245.

\bibliography{aaai2026}

\newpage
\appendix
\onecolumn
\section*{Appendix}
This appendix accompanies the paper published in the AAAI-26 proceedings and is not part of the proceedings version.
\FloatBarrier
\section{Qualitative Results}
\label{sec:qualitative}

\FloatBarrier
\subsection{Image Colorization}
\label{sec:colorization_qual}

\begin{figure}[!htbp]
  \centering
  \setlength{\tabcolsep}{2pt}
  \renewcommand{\arraystretch}{0}
  \resizebox{\textwidth}{!}{%
    \begin{tabular}{c ccc@{\hspace{10pt}}ccc}
      & & \Large Segmentation Regularized & \Large Baseline 
        & & \Large Depth Regularized    & \Large Baseline      \\
      & \Large Ground Truth           & \Large ChromaGAN + DeepLabv3 
        & \Large ChromaGAN 
        & \Large Ground Truth           & \Large ColorFormer + Midas 
        & \Large ColorFormer   \\
      \raisebox{1cm}{\rotatebox{90}{\Large Colorized outputs}}\hspace{2mm} &
      \includegraphics[height=5cm,keepaspectratio]{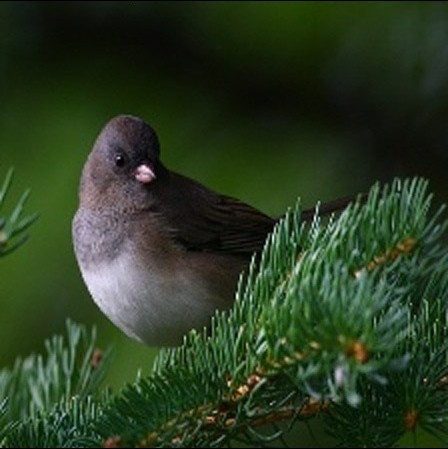}  & 
      \includegraphics[height=5cm,keepaspectratio]{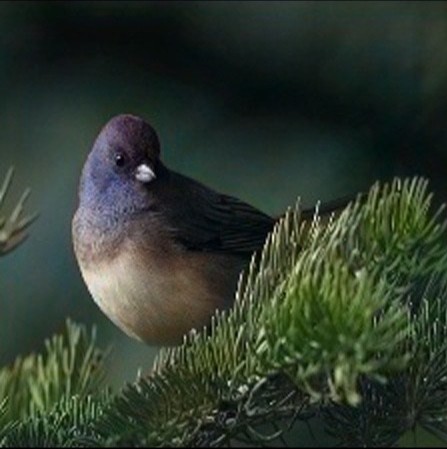}  & 
      \includegraphics[height=5cm,keepaspectratio]{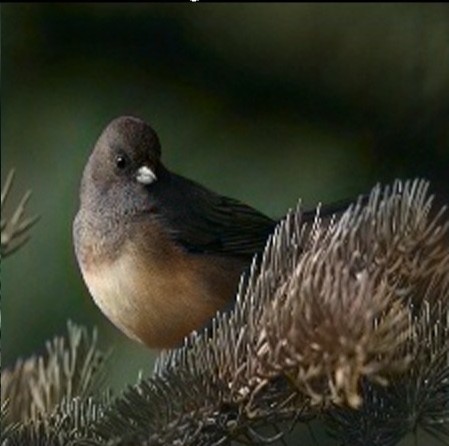}  &
      \includegraphics[height=5cm,keepaspectratio]{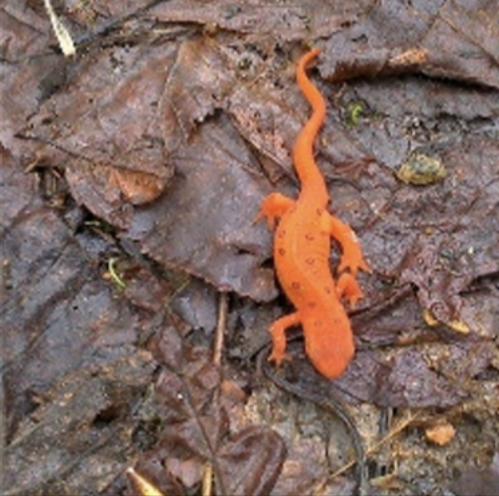}  & 
      \includegraphics[height=5cm,keepaspectratio]{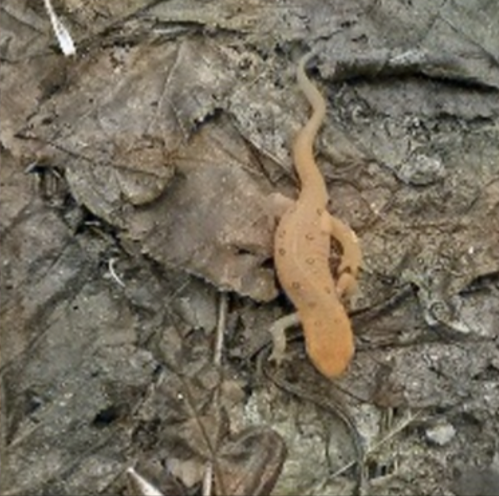}  & 
      \includegraphics[height=5cm,keepaspectratio]{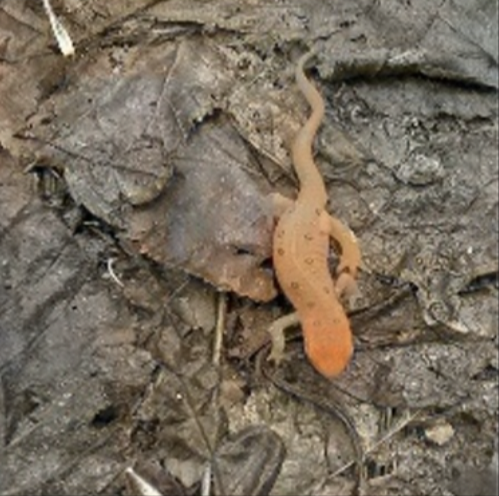}  \\
      [4pt]
      \raisebox{1cm}{\rotatebox{90}{\Large Invariant masks}}\hspace{2mm} &
      \includegraphics[height=5cm,keepaspectratio]{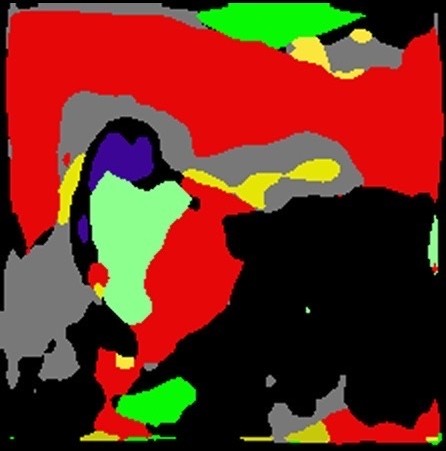}  & 
      \includegraphics[height=5cm,keepaspectratio]{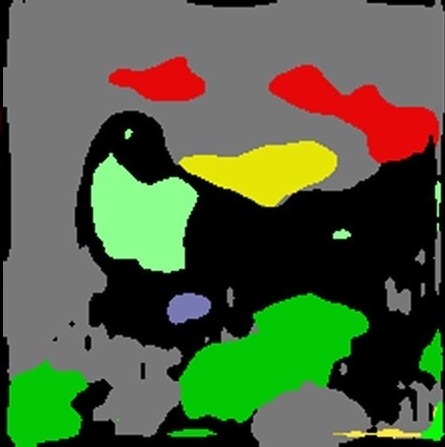}  & 
      \includegraphics[height=5cm,keepaspectratio]{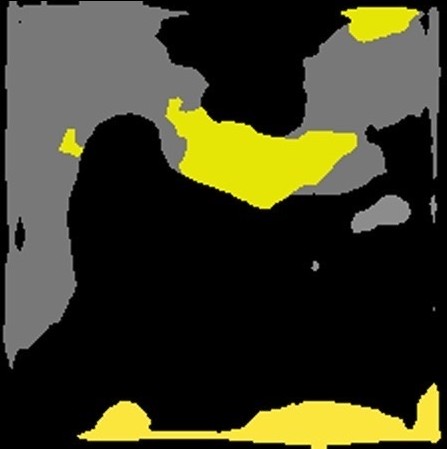}  &
      \includegraphics[height=5cm,keepaspectratio]{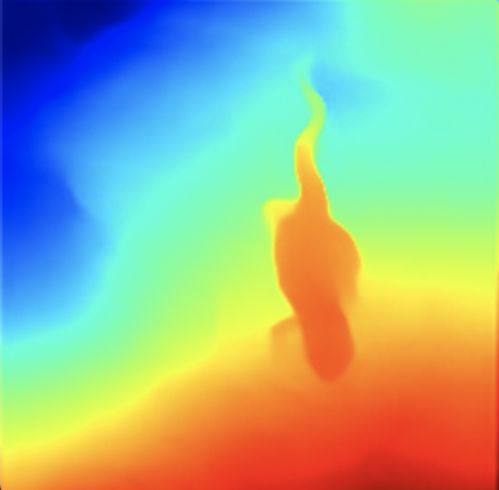} & 
      \includegraphics[height=5cm,keepaspectratio]{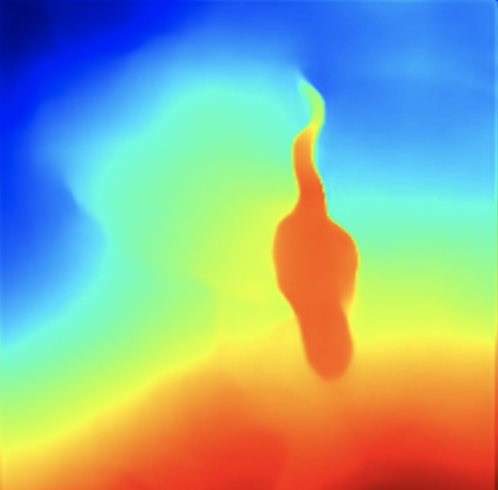} & 
      \includegraphics[height=5cm,keepaspectratio]{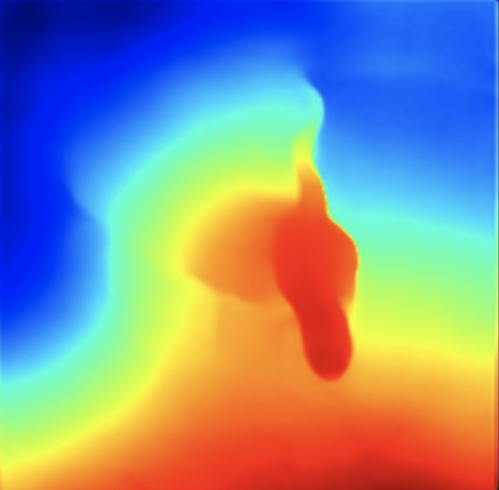} \\
    \end{tabular}%
  }

    \caption{Effectiveness of invariant-based regularization in improving image colorization. \textbf{Left panel}: Comparison using segmentation invariant. \textbf{Right panel}: Comparison using depth invariant.
Each panel shows the colorized outputs (top row) and corresponding invariant masks (bottom row). Invariant-regularized models significantly preserve critical structural details, such as object boundaries and spatial depth cues, compared to baseline models that exhibit noticeable artifacts and inaccuracies.}
  
  \label{fig:color_comparison1}
  
\end{figure}

\begin{figure}[!htbp]
  \centering
  \setlength{\tabcolsep}{2pt}
  \renewcommand{\arraystretch}{0}
  \resizebox{\textwidth}{!}{%
    \begin{tabular}{c ccc@{\hspace{10pt}}ccc}
      & & \Large Segmentation Regularized & \Large Baseline 
        & & \Large Depth Regularized    & \Large Baseline      \\
      & \Large Ground Truth           & \Large ChromaGAN + SAN
        & \Large ChromaGAN 
        & \Large Ground Truth           & \Large ColorFormer + Midas 
        & \Large ColorFormer   \\
      \raisebox{1cm}{\rotatebox{90}{\Large Colorized outputs}}\hspace{2mm} &
      \includegraphics[height=5cm,keepaspectratio]{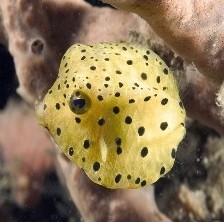}  & 
      \includegraphics[height=5cm,keepaspectratio]{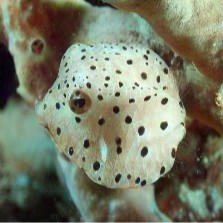}  & 
      \includegraphics[height=5cm,keepaspectratio]{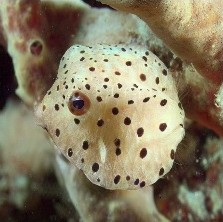}  &
      \includegraphics[height=5cm,keepaspectratio]{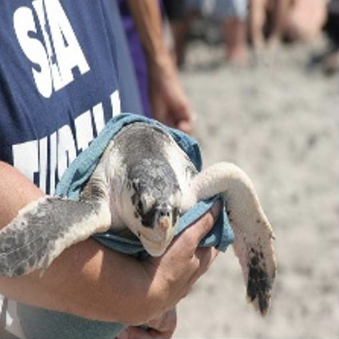}  & 
      \includegraphics[height=5cm,keepaspectratio]{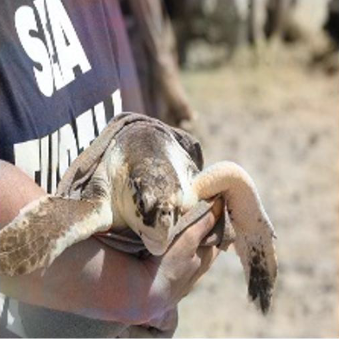}  & 
      \includegraphics[height=5cm,keepaspectratio]{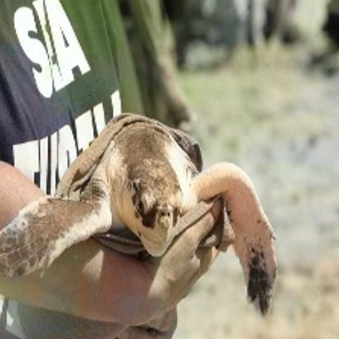}  \\
      [4pt]
      \raisebox{1cm}{\rotatebox{90}{\Large Invariant masks}}\hspace{2mm} &
      \includegraphics[height=5cm,keepaspectratio]{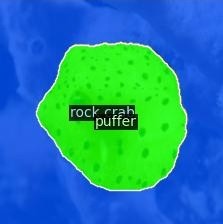}  & 
      \includegraphics[height=5cm,keepaspectratio]{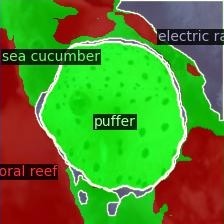}  & 
      \includegraphics[height=5cm,keepaspectratio]{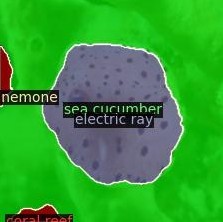}  &
      \includegraphics[height=5cm,keepaspectratio]{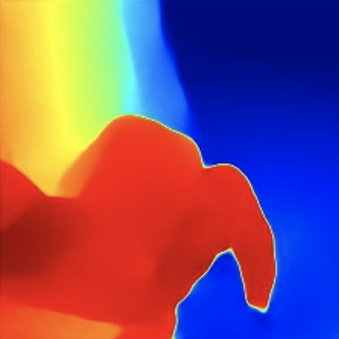} & 
      \includegraphics[height=5cm,keepaspectratio]{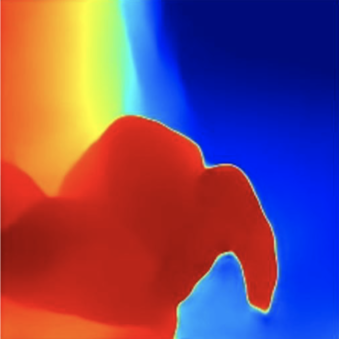} & 
      \includegraphics[height=5cm,keepaspectratio]{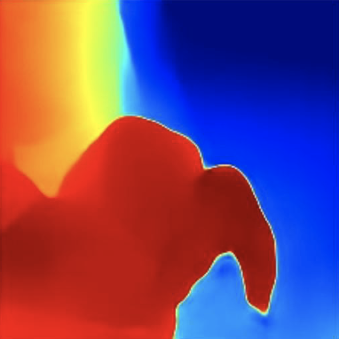} \\
    \end{tabular}%
  }

    \caption{Effectiveness of invariant-based regularization in improving image colorization. \textbf{Left panel}: Comparison using segmentation invariant. \textbf{Right panel}: Comparison using depth invariant.
Each panel shows the colorized outputs (top row) and corresponding invariant masks (bottom row). Invariant-regularized models significantly preserve critical structural details, such as object boundaries and spatial depth cues, compared to baseline models that exhibit noticeable artifacts and inaccuracies.}
  
  \label{fig:color_comparison2}
  
\end{figure}

\FloatBarrier
\clearpage
\subsection{Machine Translation}
\label{sec:translation_qual}
\begin{table}[!htbp]
\centering
\renewcommand{\arraystretch}{1.25}
\resizebox{\textwidth}{!}{%
\begin{tabular}{|l|l|}

\hline
\multicolumn{2}{|c|}{\large \textbf{Example 1}} \\
\hline
\textbf{Source Sentence:} & ``heinrich , auf der anderen seite , wird zum geburtstermin geboren , er ist also am 1 marz 1980 geboren ,'' \\ 
\textbf{Target Sentence:} & ``henry , by contrast , is born at term , so he ' s born on march 1 , 1980 .'' \\ 
\textbf{Target Logits:} & 
\begin{tabular}{|c|c|c|c|c|c|c|c|c|}
\hline
0.99992 & 0.00001 & 0.00000 & 0.00003 & 0.00001 & 0.00001 & 0.00001 & 0.00001 & 0.00001 \\ \hline
O & B-MIS & I-MIS &  B-PER &  I-PER &  B-ORG &  I-ORG &  B-LOC & I-LOC \\ \hline
\end{tabular} \\ 
\textbf{BiBERT+NER Pred Sentence:} & ``henry , on the other hand , is born at the time of birth , so he was born on march 1 , 1980 .'' \\ 
\textbf{BiBERT+NER Logits:} & 
\begin{tabular}{|c|c|c|c|c|c|c|c|c|}
\hline
0.99993 & 0.00001 & 0.00000 & 0.00002 & 0.00001 & 0.00001 & 0.00001 & 0.00001 & 0.00001 \\ \hline
O & B-MIS & I-MIS &  B-PER &  I-PER &  B-ORG &  I-ORG &  B-LOC & I-LOC \\ \hline
\end{tabular} \\ 
\textbf{BiBERT Pred Sentence:} & ``and tom , on the other hand , is born at the time of birth , so he was born on march 1 , 1980 .'' \\ 
\textbf{BiBERT Logits:} & 
\begin{tabular}{|c|c|c|c|c|c|c|c|c|}
\hline
0.99996 & 0.00000 & 0.00000 & 0.00001 & 0.00000 & 0.00000 & 0.00000 & 0.00001 & 0.00000 \\ \hline
O & B-MIS & I-MIS &  B-PER &  I-PER &  B-ORG &  I-ORG &  B-LOC & I-LOC \\ \hline
\end{tabular} \\

\hline
\multicolumn{2}{|c|}{\large \textbf{Example 2}} \\
\hline
\textbf{Source Sentence:} & ``da gab ' s die kernforschungsanlage leopoldshafen .'' \\ 
\textbf{Target Sentence:} & ``there was the leopoldshafen nuclear research plant .'' \\ 
\textbf{Target Logits:} & 
\begin{tabular}{|c|c|c|c|c|c|c|c|c|}
\hline
0.99512 & 0.00295 & 0.00220 & 0.00265 & 0.00175 & 0.00366 & 0.00625 & 0.01567 & 0.01369 \\ \hline
O & B-MIS & I-MIS &  B-PER &  I-PER &  B-ORG &  I-ORG &  B-LOC & I-LOC \\ \hline
\end{tabular} \\ 
\textbf{BiBERT+NER Pred Sentence:} & ``there was the nuclear research plant in albertz .'' \\ 
\textbf{BiBERT+NER Logits:} & 
\begin{tabular}{|c|c|c|c|c|c|c|c|c|}
\hline
0.99929 & 0.00009 & 0.00004 & 0.00006 & 0.00004 & 0.00008 & 0.00010 & 0.00015 & 0.00015 \\ \hline
O & B-MIS & I-MIS &  B-PER &  I-PER &  B-ORG &  I-ORG &  B-LOC & I-LOC \\ \hline
\end{tabular} \\ 
\textbf{BiBERT Pred Sentence:} & ``there was the nuclear research facility .'' \\ 
\textbf{BiBERT Logits:} & 
\begin{tabular}{|c|c|c|c|c|c|c|c|c|}
\hline
0.99994 & 0.00001 & 0.00000 & 0.00001 & 0.00000 & 0.00001 & 0.00001 & 0.00001 & 0.00001 \\ \hline
O & B-MIS & I-MIS &  B-PER &  I-PER &  B-ORG &  I-ORG &  B-LOC & I-LOC \\ \hline
\end{tabular} \\

\hline
\multicolumn{2}{|c|}{\large \textbf{Example 3}} \\
\hline
\textbf{Source Sentence:} & ``vor vier jahren starteten ray kurzweil und ich hier bei ted eine neue universitat , die ``singularity university'' heißt .'' \\ 
\textbf{Target Sentence:} & ``four years ago here at ted , ray kurzweil and i started a new university called singularity university .'' \\ 
\textbf{Target Logits:} & 
\begin{tabular}{|c|c|c|c|c|c|c|c|c|}
\hline
0.99928 & 0.00007 & 0.00003 & 0.00008 & 0.00004 & 0.00007 & 0.00009 & 0.00017 & 0.00017 \\ \hline
O & B-MIS & I-MIS &  B-PER &  I-PER &  B-ORG &  I-ORG &  B-LOC & I-LOC \\ \hline
\end{tabular} \\ 
\textbf{BiBERT+NER Pred Sentence:} & ``four years ago , ray kurzweil and i started a new university here at ted called singularity university .'' \\ 
\textbf{BiBERT+NER Logits:} & 
\begin{tabular}{|c|c|c|c|c|c|c|c|c|}
\hline
0.99948 & 0.00006 & 0.00003 & 0.00008 & 0.00004 & 0.00006 & 0.00010 & 0.00008 & 0.00008 \\ \hline
O & B-MIS & I-MIS &  B-PER &  I-PER &  B-ORG &  I-ORG &  B-LOC & I-LOC \\ \hline
\end{tabular} \\ 
\textbf{BiBERT Pred Sentence:} & ``four years ago , ray kurzweil and i , here at ted , started a new university called singular university .'' \\ 
\textbf{BiBERT Logits:} & 
\begin{tabular}{|c|c|c|c|c|c|c|c|c|}
\hline
0.99957 & 0.00005 & 0.00002 & 0.00009 & 0.00005 & 0.00005 & 0.00007 & 0.00007 & 0.00005 \\ \hline
O & B-MIS & I-MIS &  B-PER &  I-PER &  B-ORG &  I-ORG &  B-LOC & I-LOC \\ \hline
\end{tabular} \\

\hline
\multicolumn{2}{|c|}{\large \textbf{Example 4}} \\
\hline
\textbf{Source Sentence:} & ``er forderte octavio auf , sein stabschef zu werden .'' \\ 
\textbf{Target Sentence:} & ``he asked octavio to be his chief of staff .'' \\ 
\textbf{Target Logits:} & 
\begin{tabular}{|c|c|c|c|c|c|c|c|c|}
\hline
0.99958 & 0.00003 & 0.00002 & 0.00017 & 0.00005 & 0.00004 & 0.00004 & 0.00004 & 0.00003 \\ \hline
O & B-MIS & I-MIS &  B-PER &  I-PER &  B-ORG &  I-ORG &  B-LOC & I-LOC \\ \hline
\end{tabular} \\ 
\textbf{BiBERT+NER Pred Sentence:} & ``he asked octavio to be his chief .'' \\ 
\textbf{BiBERT+NER Logits:} & 
\begin{tabular}{|c|c|c|c|c|c|c|c|c|}
\hline
0.99926 & 0.00005 & 0.00003 & 0.00028 & 0.00010 & 0.00008 & 0.00007 & 0.00007 & 0.00006 \\ \hline
O & B-MIS & I-MIS &  B-PER &  I-PER &  B-ORG &  I-ORG &  B-LOC & I-LOC \\ \hline
\end{tabular} \\ 
\textbf{BiBERT Pred Sentence:} & ``he asked octavio to become his chief residency .'' \\ 
\textbf{BiBERT Logits:} & 
\begin{tabular}{|c|c|c|c|c|c|c|c|c|}
\hline
0.99946 & 0.00005 & 0.00003 & 0.00005 & 0.00002 & 0.00008 & 0.00006 & 0.00014 & 0.00011 \\ \hline
O & B-MIS & I-MIS &  B-PER &  I-PER &  B-ORG &  I-ORG &  B-LOC & I-LOC \\ \hline
\end{tabular} \\

\hline
\end{tabular}
} %
\caption{Comparative analysis of Machine Translation Models with and without Named Entity Recognition (NER) Invariant Regularization. Each example shows the original German source sentence, the target English translation, and predicted translations from two models: BiBERT+NER (with NER regularization) and BiBERT (baseline). The tables include detailed logit distributions across named entity categories. The results illustrate how the NER regularization significantly preserves the correct named entities and their classifications, whereas the baseline model frequently omits or misclassifies named entities, highlighting the practical advantages of invariant-based regularization.}
\label{table:ner_logits_multiple_appendix}
\end{table}

\FloatBarrier
\clearpage
\subsection{Image Captioning}
\label{sec:captioning_qual}
\begin{figure}[!htbp]
 \centering
 \begin{minipage}[t]{0.38\textwidth}
   \raisebox{-0.5\height}{\includegraphics[width=\linewidth]{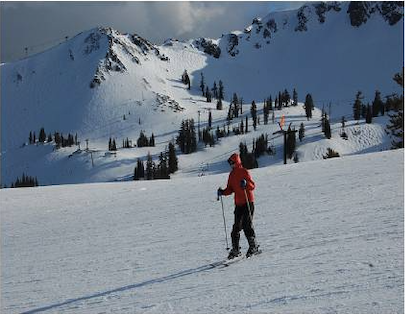}}
 \end{minipage}%
 \begin{minipage}[t]{0.62\textwidth}
   \tiny
   \setlength{\tabcolsep}{1.5pt}
   \renewcommand{\arraystretch}{0.9}
   \begin{tabular}{|p{0.30\textwidth}|p{0.30\textwidth}|p{0.30\textwidth}|}
   \hline
   \textbf{Ground Truth} & \textbf{SmallCap+NER+SceneGraph} & \textbf{SmallCap} \\ \hline
   
   \itshape Caption: A man riding skis down a snow covered slope. &
   \itshape Caption: A man riding skis down a snow covered slope. &
   \itshape Caption: A person on skis in the snow. \\ \hline
   
   \textbf{Entities:} & \textbf{Entities:} & \textbf{Entities:} \\
   \begin{tabular}[t]{@{}p{0.08\textwidth}p{0.11\textwidth}p{0.08\textwidth}@{}}
   Head & Span & Modifiers \\ \hline
   man & a man & a \\
   skis & skis & \\
   slope & a snow covered slope & a, covered \\
   \end{tabular} &
   \begin{tabular}[t]{@{}p{0.08\textwidth}p{0.11\textwidth}p{0.08\textwidth}@{}}
   Head & Span & Modifiers \\ \hline
   man & a man & a \\
   skis & skis & \\
   slope & a snow covered slope & a, covered \\
   \end{tabular} &
   \begin{tabular}[t]{@{}p{0.08\textwidth}p{0.11\textwidth}p{0.08\textwidth}@{}}
   Head & Span & Modifiers \\ \hline
   person & a person & a \\
   skis & skis & \\
   snow & the snow & the \\
   \end{tabular} \\ \hline
   
   \textbf{Relations:} & \textbf{Relations:} & \textbf{Relations:} \\
   \begin{tabular}[t]{@{}p{0.09\textwidth}p{0.09\textwidth}p{0.09\textwidth}@{}}
   Subject & Relation & Object \\ \hline
   man & riding & skis \\
   man & down & slope \\
   \end{tabular} &
   \begin{tabular}[t]{@{}p{0.09\textwidth}p{0.09\textwidth}p{0.09\textwidth}@{}}
   Subject & Relation & Object \\ \hline
   man & riding & skis \\
   man & down & slope \\
   \end{tabular} &
   \begin{tabular}[t]{@{}p{0.09\textwidth}p{0.09\textwidth}p{0.09\textwidth}@{}}
   Subject & Relation & Object \\ \hline
   person & on & skis \\
   skis & in & snow \\
   \end{tabular} \\ \hline
   \end{tabular}
 \end{minipage}
 \caption{Comparison of image captions produced by models trained with and without scenegraph and NER invariants for an image of a person skiing (left).}  
\end{figure}

\begin{figure}[!htbp]
 \centering
 \begin{minipage}[t]{0.38\textwidth}
   \raisebox{-0.5\height}{\includegraphics[width=\linewidth]{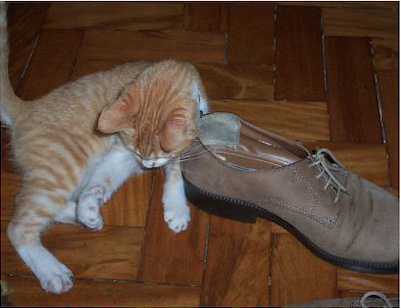}}
 \end{minipage}%
 \begin{minipage}[t]{0.62\textwidth}
   \tiny
   \setlength{\tabcolsep}{1.5pt}
   \renewcommand{\arraystretch}{0.9}
   \begin{tabular}{|p{0.30\textwidth}|p{0.30\textwidth}|p{0.30\textwidth}|}
   \hline
   \textbf{Ground Truth} & \textbf{SmallCap+NER+SceneGraph} & \textbf{SmallCap} \\ \hline
   
   \itshape Caption: An orange and white cat laying on top of a brown shoe. &
   \itshape Caption: a brown and white cat laying on top of a shoe. &
   \itshape Caption: a cat sniffing a pair of shoes on the floor. \\ \hline
   
   \textbf{Entities:} & \textbf{Entities:} & \textbf{Entities:} \\
   \begin{tabular}[t]{@{}p{0.08\textwidth}p{0.11\textwidth}p{0.08\textwidth}@{}}
   Head & Span & Modifiers \\ \hline
   cat & an orange and white cat & an, orange, white \\
   shoe & a brown shoe & a, brown \\
   \vspace*{-\baselineskip}\mbox{} & & \\
   \end{tabular} &
   \begin{tabular}[t]{@{}p{0.08\textwidth}p{0.11\textwidth}p{0.08\textwidth}@{}}
   Head & Span & Modifiers \\ \hline
   cat & a brown and white cat & a, brown, white \\
   shoe & a shoe & a \\
   \vspace*{-\baselineskip}\mbox{} & & \\
   \end{tabular} &
   \begin{tabular}[t]{@{}p{0.08\textwidth}p{0.11\textwidth}p{0.08\textwidth}@{}}
   Head & Span & Modifiers \\ \hline
   cat & a cat & a \\
   pair & a pair & a \\
   shoes & shoes & \\
   floor & the floor & the \\
   \end{tabular} \\ \hline
   
   \textbf{Relations:} & \textbf{Relations:} & \textbf{Relations:} \\
   \begin{tabular}[t]{@{}p{0.09\textwidth}p{0.09\textwidth}p{0.09\textwidth}@{}}
   Subject & Relation & Object \\ \hline
   
   \vspace*{-\baselineskip}\mbox{} & & \\
   \end{tabular} &
   \begin{tabular}[t]{@{}p{0.09\textwidth}p{0.09\textwidth}p{0.09\textwidth}@{}}
   Subject & Relation & Object \\ \hline
   
   \vspace*{-\baselineskip}\mbox{} & & \\
   \end{tabular} &
   \begin{tabular}[t]{@{}p{0.09\textwidth}p{0.09\textwidth}p{0.09\textwidth}@{}}
   Subject & Relation & Object \\ \hline
   cat & sniffing & pair \\
   pair & of & shoes \\
   cat & on & floor \\
   \end{tabular} \\ \hline
   \end{tabular}
 \end{minipage}
 \caption{Comparison of image captions produced by models trained with and without scenegraph and NER invariants for an image of a cat with a shoe (left).}  
\end{figure}

\begin{figure}[!htbp]
 \centering
 \begin{minipage}[t]{0.38\textwidth}
   \raisebox{-0.5\height}{\includegraphics[width=\linewidth]{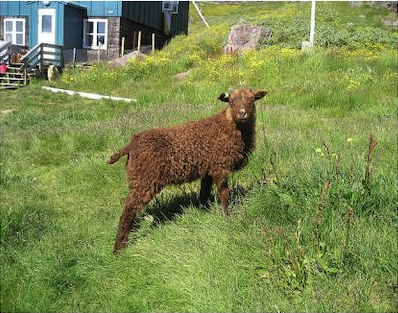}}
 \end{minipage}%
 \begin{minipage}[t]{0.62\textwidth}
   \tiny
   \setlength{\tabcolsep}{1.5pt}
   \renewcommand{\arraystretch}{0.9}
   \begin{tabular}{|p{0.30\textwidth}|p{0.30\textwidth}|p{0.30\textwidth}|}
   \hline
   \textbf{Ground Truth} & \textbf{SmallCap+NER+SceneGraph} & \textbf{SmallCap} \\ \hline
   
   \itshape Caption: A brown sheep standing on top of a lush green field. &
   \itshape Caption: a sheep standing on top of a lush green field. &
   \itshape Caption: a sheep that is standing in the grass. \\ \hline
   
   \textbf{Entities:} & \textbf{Entities:} & \textbf{Entities:} \\
   \begin{tabular}[t]{@{}p{0.08\textwidth}p{0.11\textwidth}p{0.08\textwidth}@{}}
   Head & Span & Modifiers \\ \hline
   sheep & a brown sheep & a, brown \\
   field & a lush green field & a, lush, green \\
   \vspace*{-\baselineskip}\mbox{} & & \\
   \end{tabular} &
   \begin{tabular}[t]{@{}p{0.08\textwidth}p{0.11\textwidth}p{0.08\textwidth}@{}}
   Head & Span & Modifiers \\ \hline
   sheep & a sheep & a \\
   field & a lush green field & a, lush, green \\
   \vspace*{-\baselineskip}\mbox{} & & \\
   \end{tabular} &
   \begin{tabular}[t]{@{}p{0.08\textwidth}p{0.11\textwidth}p{0.08\textwidth}@{}}
   Head & Span & Modifiers \\ \hline
   sheep & a sheep & a \\
   that & that & \\
   grass & the grass & the \\
   \end{tabular} \\ \hline
   
   \textbf{Relations:} & \textbf{Relations:} & \textbf{Relations:} \\
   \begin{tabular}[t]{@{}p{0.09\textwidth}p{0.09\textwidth}p{0.09\textwidth}@{}}
   Subject & Relation & Object \\ \hline
   
   \vspace*{-\baselineskip}\mbox{} & & \\
   \end{tabular} &
   \begin{tabular}[t]{@{}p{0.09\textwidth}p{0.09\textwidth}p{0.09\textwidth}@{}}
   Subject & Relation & Object \\ \hline

   \vspace*{-\baselineskip}\mbox{} & & \\
   \end{tabular} &
   \begin{tabular}[t]{@{}p{0.09\textwidth}p{0.09\textwidth}p{0.09\textwidth}@{}}
   Subject & Relation & Object \\ \hline
   that & in & grass \\
   \vspace*{-\baselineskip}\mbox{} & & \\
   \end{tabular} \\ \hline
   \end{tabular}
 \end{minipage}
 \caption{Comparison of image captions produced by models trained with and without scenegraph and NER invariants for an image of a sheep in a field (left).}  
\end{figure}

\FloatBarrier
\section{Differential Performance of FBD Variants Across Tasks}
\label{sec:fbd_variants}
Our experiments reveal an intriguing phenomenon: FBD-4 computed using token averaging performs well for machine translation but poorly for image captioning, while FBD-4 computed using CLS tokens shows the opposite pattern. This divergent behavior can be attributed to fundamental differences in how BERT encodes linguistic information for these distinct tasks.

\FloatBarrier
\subsection{Task-Specific Information Distribution in BERT Layers}

\textbf{Machine Translation:} The translation task requires preserving fine-grained linguistic features across languages. Lower and middle BERT layers capture syntactic structures and local dependencies that are crucial for translation quality. The average of all tokens across the last 4 layers effectively captures both syntactic patterns (from earlier layers) and semantic coherence (from later layers). Translation necessitates maintaining word order, grammatical structures, and morphological variations, which are distributed across multiple token representations.

\textbf{Image Captioning:} In contrast, image captioning requires mapping visual concepts to linguistic descriptions, making global semantic representation more important than fine-grained syntactic details. The CLS token in later layers aggregates scene-level semantic information, which aligns better with the holistic nature of image-to-text mapping. Caption generation is less dependent on preserving exact syntactic structures compared to translation.

\FloatBarrier
\subsection{Information Aggregation Patterns}

The effectiveness of different aggregation strategies reflects the nature of information preservation required by each task:

\textbf{Token-Average (FBD-4) for Machine Translation:}
\begin{itemize}
    \item Averaging all tokens preserves local context and word-level nuances critical for translation
    \item This approach maintains information about positional relationships and local dependencies
    \item The invariants in translation (named entities, sentiment) are often localized to specific tokens, making token-level averaging more sensitive to these changes
\end{itemize}

\textbf{CLS-Token (FBD-4-CLS) for Image Captioning:}
\begin{itemize}
    \item The CLS token acts as a global sentence representation, capturing overall semantic content
    \item For captions, the relationship between visual content and textual description is more about semantic correspondence than token-level precision
    \item Scene graphs and entity recognition in images map better to global semantic representations than to individual token positions
\end{itemize}

\FloatBarrier
\subsection{Invariant Sensitivity Mechanisms}

The effectiveness of each metric variant correlates with how invariants manifest in each task. Translation invariants (NER, sentiment) are embedded within specific token sequences, making token-averaging more sensitive to their preservation. In contrast, caption invariants (scene graphs, visual entities) are better captured by global representations that abstract away from specific token positions.

\FloatBarrier
\subsection{Theoretical Formulation}

The differential performance of FBD variants can be formalized as follows:

For machine translation:
\begin{equation}
\text{FBD-4}_{\text{avg}} = D\left(\mu\left(h^{L-3:L}_{1:n}\right), \mu\left(h'^{L-3:L}_{1:n}\right)\right)
\end{equation}

For image captioning:
\begin{equation}
\text{FBD-4}_{\text{cls}} = D\left(h^{L-3:L}_{\text{CLS}}, h'^{L-3:L}_{\text{CLS}}\right)
\end{equation}

where $h$ represents hidden states, $L$ is the last layer, $n$ is sequence length, and $D$ is the Fréchet distance.

The superior performance of token-averaging for translation reflects the need to preserve fine-grained linguistic features, while CLS-based aggregation excels for captioning due to its alignment with global semantic mapping from visual to textual domains.

\FloatBarrier
\subsection{Implications for Metric Design}

This phenomenon aligns with recent findings in the literature. Tenney et al. (2019) showed that BERT's linguistic knowledge follows a classical NLP pipeline, with syntactic information in middle layers and semantic information in upper layers. Rogers et al. (2020) demonstrated that CLS tokens in final layers best capture sentence-level semantic similarity.

Our findings highlight a crucial insight: \textbf{the optimal aggregation strategy for distributional metrics depends on the task's information requirements}. This task-specific optimization of distributional metrics underscores the importance of considering the underlying information distribution when designing evaluation metrics for multi-output structured prediction tasks. Rather than applying metrics uniformly across tasks, researchers should carefully consider how information is encoded and what aspects of the output are most critical for task performance.

This observation has broader implications for the field: as we develop more sophisticated evaluation metrics, we must move beyond one-size-fits-all approaches and design metrics that align with the specific information patterns and invariant structures of each task.

\FloatBarrier
\section{Experimental Details}
All the training parameters of baseline model were kept same including learning rate. 
\label{sec:experimental}

\FloatBarrier
\subsection{Image Colorization}
\label{sec:colorization_exp}

\subsubsection{General Colorization Training Details}
\label{app:color}
\begin{itemize}
    \item ChromaGAN
    \begin{itemize}
        \item Input image size:  224×224
        \item We reimplemented ChromaGAN in PyTorch to ensure fair comparison with our PyTorch-based models, achieving identical validation performance (FID = 5.81 on 50k images) to the original TensorFlow pretrained model.
        \item All ChromaGAN results reported in our experiments correspond to the original TensorFlow pretrained model, while TaskNet models are implemented in PyTorch.
        \item Our PyTorch implementation maintains architectural fidelity and parameter count consistency with the original ChromaGAN design.

    \end{itemize}

    \item ColorFormer
    \begin{itemize}
        \item Input image size:  256×256
        \item We used same pytorch implementation of basicSR repository and used by ColorFormer
        \item We added invariant loss files in train\_colorformer.yml file 
    \end{itemize}

    \item DDColor
    \begin{itemize}
        \item Input image size:  256×256
        \item We used same pytorch implementation of basicSR repository and used by DDColor
        \item We added invariant loss files in train\_colorformer.yml file
        \item We used DDColor-tiny model for all our experiments and retrained the model. 
    \end{itemize}

\begin{table}[!htbp]
\centering
\small
\caption{Hyperparameter $\alpha$ Settings for Image Colorization Experiments. $\alpha$ represents the multiplier to invariant loss added to total training loss during colorization training.}
\label{tab:hyperparams_colorization}
\begin{tabular}{lccccc}
\toprule
\textbf{Model} & \textbf{Dataset} & \textbf{Best $\alpha$ } & \textbf{Batch Size} & \textbf{Epochs / Iterations} & \textbf{Hyperparameters Tried ($\alpha$)} \\
\midrule
ChromaGAN & ImageNet & - & 10 & 5 epochs & - \\
ChromaGAN + Seg & ImageNet & 0.02 & 10 & 15 epochs & 0.0002, 0.002, 0.02, 0.05, 0.2 \\
ChromaGAN + Depth & ImageNet & 0.0001 & 10 & 15 epochs & 0.0001, 0.001, 0.005, 0.01 \\
ChromaGAN + Seg + Depth & ImageNet & 0.02,0.0001 & 10 & 15 epochs & (0.02,0.0001),(0.02,0.005),(0.01,0.001) \\
\midrule
ColorFormer & ImageNet & - & 16 & 200,000 iterations & - \\
ColorFormer + Seg & ImageNet & 0.05 & 16 & 600,000 iterations & 0.0001, 0.001, 0.01, 0.05, 0.1, 0.2 \\
ColorFormer + Depth & ImageNet & 0.01 & 16 & 600,000 iterations & 0.0001, 0.001, 0.01, 0.05, 0.1, 0.2 \\
ColorFormer + Seg + Depth & ImageNet & 0.05,0.01 & 16 & 600,000 iterations &  (0.05,0.01),(0.01,0.001),(0.05,0.001) \\
\midrule
DDColor & ImageNet & - & 16 & 400,000 iterations & - \\
DDColor + Seg & ImageNet & 0.5 & 16 & 800,000 iterations & 0.05, 0.1, 0.5, 2, 5  \\
DDColor + Depth & ImageNet & 0.02 & 16 & 800,000 iterations & 0.001, 0.01, 0.02, 0.05, 0.1 \\
DDColor + Seg + Depth & ImageNet & 2,0.02 & 16 & 800,000 iterations & (2,0.02),(0.5,0.02),(0.1,0.01) \\
\bottomrule
\end{tabular}
\end{table}

\end{itemize}

\subsubsection{Invariant Network and Loss Details}
\begin{itemize}
    \item \textbf{Side Adapter Network:} It is an open vocabulary semantic segmentor network, which segments images based on the vocabulary given to it. For enhanced segmentation, ImageNet classes were incorporated, resulting in a total of 1171 classes. Given the predicted output \texttt{pred} and the ground truth \texttt{gt}, the Dice Loss for multi-class segmentation is calculated using the following formula:

    \begin{equation}
    \text{DiceLoss}(\texttt{pred}, \texttt{gt}) = 1 - \frac{2 \times \sum_{i=1}^{C} p_i \cdot g_i + \epsilon}{\sum_{i=1}^{C} p_i^2 + \sum_{i=1}^{C} g_i^2 + \epsilon}
    \end{equation}
    where $C$ represents the total number of classes (1171 in this case), $p_i(\texttt{pred})$ and $g_i(\texttt{gt})$ denote the predicted and ground truth probabilities for the $i^{th}$ class, respectively, and $\epsilon$ is a small constant added for numerical stability.

    \item \textbf{MiDaS:} We used the "MiDaS small" version from the PyTorch hub. MiDaS takes an RGB image as input, so during training, we convert the colorizer's Lab output to RGB. The loss function used is the silog loss.

    Let $M$ be the number of pixels in the image. Let $d = d\left(\theta\right) \in \mathbb{R}^M$ be the disparity prediction output by the depth estimator and $d = d* \in \mathbb{R}^M$ be the ground truth disparity. Then the silog loss can be written as:
    \begin{equation}
    L_{silog}\left(z,z*\right) = \max_{\mathbf{s}}\frac{1}{2 M}\sum^{M}_{i=1}\left( \log\left( e^s z_i\right) - \log\left(z*_i\right) \right)
    \end{equation}
\end{itemize}

\FloatBarrier
\subsection{Machine Translation}
\label{sec:translation_exp}

\subsubsection{General Machine Translation Training Details}
\begin{itemize}
    \item \textbf{BiBERT}
    \begin{itemize}
        \item Model architecture: BERT-based bidirectional encoder with cross-attention
        \item Hidden dimension: 768, Number of layers: 12
        \item Vocabulary size: 37,000 shared BPE tokens
        \item Maximum sequence length: 150 tokens
        \item We used the official BiBERT implementation with PyTorch
        \item Training conducted on 1 NVIDIA V100 GPUs with mixed precision
        \item Gradient accumulation steps: 4
        \item Warmup steps: 4,000
    \end{itemize}
    
    \item \textbf{Bi-SimCut}
    \begin{itemize}
        \item Model architecture: Transformer with simple data augmentation
        \item Hidden dimension: 512, Number of layers: 6
        \item Uses bidirectional training with cut-off augmentation
        \item Dropout rate: 0.3
        \item Label smoothing: 0.1
        \item Training conducted on 1 NVIDIA V100 GPUs
    \end{itemize}
    
    \begin{table}[!htbp]
\centering
\caption{Hyperparameter $\alpha$ Settings for Neural Machine Translation Experiments}
\begin{tabular}{lccccc}
\toprule
Model & Dataset & Best $\alpha$ & Batch Size & Epochs & Hyperparameters Tried ($\alpha$) \\
\midrule
BiBERT & IWSLT-14 & - & 4096 tokens & 75 & - \\
BiBERT + NER & IWSLT-14 & 100 & 4096 tokens & 75 & 10, 50, 100, 500, 1000 \\
BiBERT + SeAM & IWSLT-14 & 50 & 4096 tokens & 75 & 10, 50, 100, 500, 1000 \\
BiBERT + NER + SeAM & IWSLT-14 & 100, 50 & 4096 tokens & 75 & (100,50), (50,50), (100,100) \\
\midrule
Bi-SimCut & IWSLT-14 & - & 4096 tokens & 75 & - \\
Bi-SimCut + NER & IWSLT-14 & 200 & 4096 tokens & 75 & 50, 100, 200, 500, 1000 \\
Bi-SimCut + SeAM & IWSLT-14 & 100 & 4096 tokens & 75 & 50, 100, 200, 500, 1000 \\
Bi-SimCut + NER + SeAM & IWSLT-14 & 200, 100 & 4096 tokens & 75 & (100,50), (50,50), (200,100) \\
\bottomrule
\end{tabular}
\end{table}

\end{itemize}

\subsubsection{Invariant Network and Loss Details}
\begin{itemize}
    \item \textbf{NER (Named Entity Recognition) Network:} We use Hugging Face's \texttt{dslim/bert-large-NER} model which identifies four entity types: Person (PER), Organization (ORG), Location (LOC), and None (O). The NER loss is computed as KL divergence between predicted and ground truth entity distributions:
    
    \begin{equation}
    L_{NER} = \sum_{i=1}^{N} \sum_{c \in \{PER, ORG, LOC, O\}} p_i^c \log\left(\frac{p_i^c}{q_i^c}\right)
    \end{equation}
    
    where $N$ is the sequence length, $p_i^c$ is the ground truth probability for class $c$ at position $i$, and $q_i^c$ is the predicted probability.
    
    \item \textbf{Sentiment Analysis (SeAM) Network:} We use Hugging Face's distilbert-base-uncased-finetuned-sst-2-english for binary sentiment classification. The sentiment loss ensures translations preserve the emotional tone:
    
    \begin{equation}
    L_{SeAM} = D_{KL}(P_{sent}^{src} || P_{sent}^{tgt})
    \end{equation}
    
    where $P_{sent}^{src}$ and $P_{sent}^{tgt}$ are the sentiment distributions for source and translated sentences respectively.
\end{itemize}

\subsubsection{Training Procedure}
\begin{itemize}
    \item All models trained for 75 epochs on IWSLT-14 and 300k steps on WMT-14
    \item Learning rate schedule: inverse square root with 8,000 warmup steps
    \item Optimizer: Adam with $\beta_1=0.9$, $\beta_2=0.98$
    \item Batch size: 4,096 tokens per GPU
    \item Checkpoint selection based on validation BLEU score
    \item For invariant models, checkpoint selection based on validation FBD score
\end{itemize}

\FloatBarrier
\subsection{Image Captioning}
\label{sec:captioning_exp}

\subsubsection{General Image Captioning Training Details}
\begin{itemize}
    \item \textbf{SmallCap}
    \begin{itemize}
        \item Architecture: Lightweight transformer with retrieval augmentation
        \item Vision encoder: CLIP ViT-B/32
        \item Text decoder: GPT-2 small (117M parameters)
        \item Uses retrieval from training set to provide relevant examples
        \item Maximum caption length: 40 tokens
        \item Beam size: 5 for inference
    \end{itemize}
    
    \item \textbf{ClipCap}
    \begin{itemize}
        \item Architecture: CLIP vision encoder with GPT-2 language model
        \item Mapping network: 8-layer transformer to project CLIP features
        \item Text decoder: GPT-2 medium (345M parameters)
        \item Prefix length: 10 tokens
        \item Does not use additional retrieval mechanisms
    \end{itemize}
    
    \begin{table}[!htbp]
\centering
\caption{Hyperparameter $\alpha$ Settings for Image Captioning Experiments}
\begin{tabular}{lccccc}
\toprule
Model & Dataset & Best $\alpha$ & Batch Size & Epochs & Hyperparameters Tried ($\alpha$) \\
\midrule
SmallCap & COCO & - & 32 & 10 & - \\
SmallCap + NER & COCO & 0.5 & 32 & 10 & 0.1, 0.3, 0.5, 0.7, 1.0 \\
SmallCap + SceneGraph & COCO & 1.0 & 32 & 10 & 0.5, 1.0, 2.0, 5.0 \\
SmallCap + NER + SG & COCO & 0.5, 1.0 & 32 & 10 & (0.5,1.0), (0.3,1.0), (0.5,2.0) \\
\midrule
ClipCap & COCO & - & 64 & 10 & - \\
ClipCap + NER & COCO & 0.3 & 64 & 10 & 0.1, 0.3, 0.5, 0.7, 1.0 \\
ClipCap + SceneGraph & COCO & 0.7 & 64 & 10 & 0.5, 0.7, 1.0, 2.0 \\
ClipCap + NER + SG & COCO & 0.3, 0.7 & 64 & 10 & (0.3,0.7), (0.3,1.0), (0.5,0.7) \\
\bottomrule
\end{tabular}
\end{table}

\end{itemize}

\subsubsection{Invariant Network and Loss Details}
\begin{itemize}
    \item \textbf{Named Entity Recognition (NER):} Same BERT-large-NER model as in MT, adapted for caption text. Ensures proper nouns and entities in images are correctly identified in captions.
    
    \item \textbf{Scene Graph Parser:} Uses the Stanford Scene Graph Parser to extract structured representations. The scene graph loss compares entities and relationships:
    
    \begin{equation}
    L_{SG} = \lambda_e L_{entities} + \lambda_r L_{relations}
    \end{equation}
    
    where:
    \begin{equation}
    L_{entities} = \frac{|E_{gt} \triangle E_{pred}|}{|E_{gt} \cup E_{pred}|}
    \end{equation}
    
    \begin{equation}
    L_{relations} = \frac{|R_{gt} \triangle R_{pred}|}{|R_{gt} \cup R_{pred}|}
    \end{equation}
    
    $E_{gt}, E_{pred}$ are ground truth and predicted entity sets, $R_{gt}, R_{pred}$ are relation sets, and $\triangle$ denotes symmetric difference.
\end{itemize}

\subsubsection{Training Procedure}
\begin{itemize}
    \item Trained for 10 epochs on COCO and 15 epochs on Flickr30k
    \item Learning rate: 5e-5 with cosine annealing
    \item Optimizer: AdamW with weight decay 0.01
    \item Batch size: 32 for SmallCap, 64 for ClipCap
    \item Teacher forcing ratio: 1.0 (always use ground truth during training)
    \item Gradient clipping: max norm 1.0
\end{itemize}

\FloatBarrier
\end{document}